\documentclass{article}

\usepackage{iclr2027_conference,times}

\usepackage{hyperref}
\usepackage{url}
\usepackage{graphicx}
\usepackage{booktabs}
\usepackage{amsmath}
\usepackage{amssymb}
\usepackage{xcolor}
\usepackage{multirow}
\usepackage{array}
\usepackage[utf8]{inputenc}
\DeclareUnicodeCharacter{2212}{\textminus}
\usepackage{microtype}

\title{Lost with a Map: Conversational State and Behavioral Reliability in Language Models}

\author{Atahan Dokme \\
Georgia Institute of Technology \\
\texttt{adokme3@gatech.edu}
\And
Larry Heck \\
Georgia Institute of Technology \\
\texttt{larryheck@gatech.edu} 
}

\iclrfinalcopy

\begin{document}

\maketitle
\fancyhead{}
\renewcommand{\headrulewidth}{0pt}

\begin{abstract}
Task-oriented dialogue requires maintaining and updating information across turns,
yet language models expose no explicit belief-state object. We study how conversational state is represented, updated, and used inside eight instruction-tuned language models from four families on MultiWOZ and SGD. Structure and values separate: which domains, slots, and requests are active is linearly readable just before the model acts, whereas exact values are far more readable where the user stated them. After a user changes a value, both values remain accessible at their mentions, and causal interventions show that both continue to influence the model's action. In natural closed-loop interaction, query failures separate into cases of weak structural support, incorrect value resolution, and failure to deploy otherwise-supported constraints, with targeted interventions producing systematically different repair behavior across these cases. These findings motivate a state–action controller that starts from the base
model action and selectively edits it using structural readouts, without requiring a
complete predicted belief state as an intermediate representation. On held-out MultiWOZ interaction across five models, it raises the base model exact-query accuracy from .318 to .621 and task success from .272 to .371 at negligible added cost. Overall, reliable interaction requires not only retaining conversational information, but resolving which available constraints currently apply and ensuring that they govern action.

\end{abstract}

\section{Introduction}
\label{sec:introduction}

Multi-turn conversation requires more than responding to the latest
utterance. A system must retain earlier information, revise it when the
user's goals change, and ensure that the currently relevant information
governs subsequent actions. Classical task-oriented dialogue (TOD) systems
make this requirement explicit through a belief state that is updated before
database interaction and response generation
\citep{budzianowski-etal-2018-multiwoz,wu-etal-2019-transferable,
hosseini-asl-etal-2020-simple,yang-etal-2021-ubar,
peng-etal-2021-soloist,su-etal-2022-multi}.
General-purpose language models can instead act directly from dialogue
history without exposing an analogous state object. This raises a basic
question:

\begin{center}
\textbf{How is conversational state represented, updated, and used without an explicit belief state?}
\end{center}

We study this question across MultiWOZ and SGD using eight instruction-tuned
language models from four model families. We find a consistent separation between conversational structure and exact values: active domains, constrained slots,
and requests are readily accessible at the decision site (the final context position before the model generates its action), whereas exact values remain much more accessible at the earlier dialogue positions where they were introduced. Direct source interventions
further show that earlier information can remain causally consequential after it is superseded, while its downstream impact changes with whether the information is still current. Conversational state in language models is therefore not a clean turn-by-turn overwrite: preceding and updated constraints can remain accessible and causally relevant, so reliable behavior depends on selecting which information currently applies and ensuring that it governs the action. We refer to this pattern as \emph{append--select--deploy}: earlier information accumulates rather than being overwritten (\emph{append}), the model must resolve which of the available constraints currently applies (\emph{select}), and the resolved constraint must actually reach the emitted action (\emph{deploy}). 

In this sense the model gets \textbf{lost with a map}: the relevant conversational information is often still accessible and causally active, but the model does not reliably select and deploy it. 

In natural closed-loop interaction, query errors separate into cases of weak structural support,
incorrect value resolution, and failure to deploy otherwise-supported
constraints, and these cases respond differently to targeted interventions.
Conversational failure therefore cannot be reduced to information loss alone.
Finally, we demonstrate the measured signals can improve behavior without reinstating a complete belief-state pipeline. Our \textbf{State--Action
Controller} (SAC) starts from the Base LM query, uses a structural readout to propose targeted edits, resolves selected values from visible dialogue
evidence, and protects Base LM constraints that remain supported. Conservative, intermediate, and binary operating points intervene progressively more often. On held-out MultiWOZ interaction, SAC improves task success and query fidelity over the Base LM and prompted state reconstruction. Compared with a supervision-matched T5 tracker, SAC attains equivalent mean terminal success and repair while producing more faithful queries, preserving correct Base LM actions substantially more reliably, and making approximately nine times fewer destructive corrections (Section~\ref{sec:control-test}).

\paragraph{Contributions.}
First, across eight models and two datasets, we characterize the representational substrate for conversational state: structural information is accessible near the decision site, while exact values remain strongly accessible at their historical mentions. Second, using matched causal interventions, we show that historical sources remain behaviorally consequential after updates, and that their influence changes systematically with conversational validity. Third, we connect these measurements to natural closed-loop failures: omitted constraints exhibit distinct operational structure, binding, and deployment profiles that predict different responses to targeted intervention. Finally, we show that these interpretations and measurements can support selective action correction through a State–Action Controller, without requiring complete state reconstruction on every inference path.

\section{Related Work}
\label{sec:related}

\paragraph{Dialogue state tracking (DST) and language-model-based TOD.}
Task-oriented dialogue has traditionally placed an explicit belief state
between dialogue understanding and downstream database interaction
\citep{budzianowski-etal-2018-multiwoz,wu-etal-2019-transferable}.
Subsequent systems increasingly unified state tracking, policy, and response
generation within pretrained language models
\citep{hosseini-asl-etal-2020-simple,yang-etal-2021-ubar,
peng-etal-2021-soloist,su-etal-2022-multi}, while specialized trackers use
copying and selective state-update mechanisms
\citep{heck-etal-2020-trippy}. More recent work elicits explicit slot--value
state directly from large language models through prompting or adaptation
\citep{hu-etal-2022-context,heck-etal-2023-chatgpt,
feng-etal-2023-towards}. Our setting differs: the model already produces
executable actions directly from dialogue history, and we use latent state
information to selectively correct those actions rather than reconstructing a
complete belief state on every inference path.

\paragraph{Internal state, historical information, and latent control.}
Prior probing work shows that reasoning states, dialogue, and task variables can be decoded from
pretrained representations \citep{dokme2026selectivestatespaceadaptationretrieval, wu-xiong-2020-probing,wu-etal-2020-tod},  while studies of entity binding, tracking, and retrieval examine how language models recover information tied to earlier mentions
\citep{ICLR2025_9b77f073,feng2023binding,kim2023entity,
prakash2024finetuning,prakash2026lookbacks,oh2026rebinding}.
Because decodability does not establish causal or behavioral use
\citep{elazar-etal-2021-amnesic,ravfogel-etal-2021-counterfactual,
lasri-etal-2022-probing}, our matched interventions hold the source and value
fixed while changing whether that information remains conversationally valid. Probes and activation differences have also been used to steer or monitor model behavior
\citep{avsian2026latentimlatentinteractionmanagement, li2023inferencetime,rimsky-etal-2024-steering,azaria-mitchell-2023-internal},  while constrained decoding and verification operate on emitted programs \citep{scholak2021picardparsingincrementallyconstrained,  wang2018robusttexttosqlgenerationexecutionguided, pmlr-v202-ni23b}.
Our controller connects these directions by using a latent structural readout
to selectively edit an executable action rather than modifying activations or
reconstructing a complete belief state. This complements work on degradation over extended interaction \citep{laban-etal-2026-lost,tack-etal-2026-evolving} showing that models lose track of user intent over multi-turn interaction; our results indicate that relevant information often remains accessible and causally active, with failures arising in selecting and deploying it.

\section{Conversational State: Setting and Accessibility}
\label{sec:state}

\subsection{Setting and Measurement Framework}
\label{sec:state-setting}

At turn $t$, let $H_t$ denote the dialogue history and let the unmodified
language-model policy produce an executable query $q_t = f_\theta(H_t)$. For
analysis, the reference is the annotated current conversational state: the active
domain, the service intent (in SGD), the
constrained slots, the requested information, and the corresponding exact
values. Our main question is how this current state is represented when no
explicit belief-state object lies on the inference path. For an update
$(s,v_{\mathrm{old}}) \rightarrow (s,v_{\mathrm{new}})$, we call the mention
supporting $v_{\mathrm{new}}$ the \emph{current-value source} and the earlier
mention supporting $v_{\mathrm{old}}$ the \emph{superseded-value source}.
These labels describe conversational validity, not internal accessibility: a
value may cease to be correct while remaining represented where it originally
entered the conversation. We evaluate eight instruction-tuned models from four model families on
MultiWOZ~\citep{budzianowski-etal-2018-multiwoz,zang-etal-2020-multiwoz} and SGD~\citep{rastogi2020scalablemultidomainconversationalagents}, named by family and size (e.g., Llama-8B-IT; exact checkpoints in Table~\ref{tab:model-inventory}). We distinguish the \emph{decision site} (SEARCH-PRE), the
final context position before the relevant generation, from
\emph{mention sites} where values entered the dialogue. Readouts are fit on
dialogue-disjoint splits and frozen before evaluation; full fitting,
calibration, and preprocessing details appear in
Appendix~\ref{app:readouts}.

\subsection{Structural Information and Exact Value Accessibility}
\label{sec:state-structure}

On MultiWOZ, linear readouts at the decision site recover the structural
components of conversational state reliably across
all eight models: domain F1 ranges from $.926$ to $.976$, constrained-slot F1
from $.752$ to $.808$, and request F1 from $.662$ to $.802$. Thus, immediately before generation, substantial information about
\emph{which} parts of the conversational state are active is linearly
accessible, and part of it is recoverable through a rank-32 supervised
projection that outperforms rank-matched PCA and shuffled-supervision
controls in every model--task combination (Figure~\ref{fig:accessibility}a;
Appendix~\ref{app:structural-readouts}).

Exact values exhibit a different localization pattern. To compare locations
without changing the evaluation population, we join mention-site and
decision-site records by dialogue, slot, and value, and evaluate both frozen
readouts on the same $3{,}255$ held-out instances. In every instance, the
decision site occurs exactly three turns after the historical value mention.
Across the eight models, exact-value identity is recovered from the
mean-pooled mention-span representation with accuracy $.961$--$.993$, but
only $.182$--$.313$ from the decision-site representation. The paired
mention--decision-site gap ranges from $+.672$ to $+.808$, with every
dialogue-level bootstrap confidence interval excluding zero
(Figure~\ref{fig:accessibility}b; Appendix~\ref{app:value-localization}). This gap is not a lack of the value itself: when the slot key is
teacher-forced immediately after the decision site (domain: \{domain\}; \{slot\}:), the same probe recovers the value at the key position with
accuracy $.618$--$.942$ on all eight checkpoints, against
$.182$--$.313$ one step earlier, so values become linearly readable once
the key is chosen while names and times remain mention-bound
(Appendix~\ref{app:value-localization}).

\subsection{Earlier and Updated Values Remain Accessible}
\label{sec:state-updates}

SGD slot-update events change a slot from
$v_{\mathrm{old}}$ to $v_{\mathrm{new}}$. 
Two questions follow: after the update, is the old value still readable at its original mention, and can a decision-site readout identify which of the two values is now current? Across $304$ events from $258$
dialogues, the old value remains readable at its original mention with
accuracy $.868$--$.898$ across models, while the new value is readable at its
own mention with accuracy $.885$--$.941$. Thus, earlier and updated values can
remain simultaneously accessible after an update. Each readout is a forced choice between the event's two candidate values, so chance accuracy is .5.

\begin{figure}[!t]
    \centering
    \includegraphics[width=0.99\textwidth]{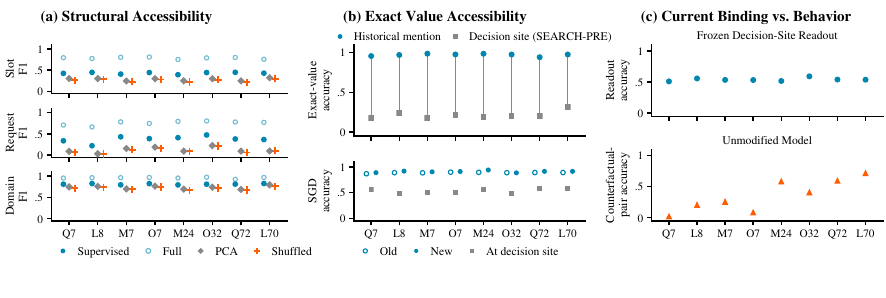}
    \caption{\textbf{Accessibility of conversational state across eight models.}
    Model abbreviations are defined in Table~\ref{tab:model-inventory}.
    (a) Rank-32 supervised projections recover structural variables at the
    decision site better than rank-matched controls.
    (b) Exact values are more readable at their historical mentions than at the
    decision site exactly three turns later on the same $3{,}255$ instances;
    SGD updates similarly preserve both old and new values at their mentions,
    while current-binding readout remains near chance.
    (c) On matched counterfactual dialogues, frozen decision-site binding
    readout is near chance, whereas native behavior improves with scale.
    The two strips use different units: item-level readout accuracy above and
    pair-level behavioral accuracy below.}
    \label{fig:accessibility}
    \end{figure}

Current binding (identifying which of two competing values for a slot
currently applies) is less exposed through the same interface. A frozen
decision-site readout picks the currently valid member of an old/new pair
with only $.510$--$.591$ accuracy on controlled MultiWOZ-derived counterfactual pairs
(Figure~\ref{fig:accessibility}c) and $.477$--$.576$ on SGD. The models
themselves do better with scale: the counterfactual pairs share two candidate
values and differ only in a final cue, and a pair counts as correct only if
the unmodified model selects the valid value in both members. Pair accuracy
ranges from $.027$ (Qwen-7B-IT) to $.721$ (Llama-70B-IT),
and on items the readout gets wrong, Mistral-24B-IT and the two 70B-class
checkpoints still choose correctly $.76$--$.82$ of the time
(Appendix~\ref{app:readout-behavior-disagreement}). Weak readout accuracy
therefore does not imply absent competence; the tested interface fails to
expose it, and nonlinear readouts under the same protocol do no better
($.494$--$.559$ against $.510$--$.557$ linear on four checkpoints;
Appendix~\ref{app:current-binding-readouts}).

These mention-side asymmetries are in parallel with the causal transformer architecture: a value's representation at its mention position cannot be revised by later tokens. However, the decision-site half is not: nothing prevents the model from consolidating the current value of each slot at the position where it acts, and the readouts above recover structure there but not values. Thus, the contribution of this section is that these asymmetries set up a state-resolution problem. Classical task-oriented systems expose an explicit current belief state that is updated before downstream action. The language models studied here instead act directly over an accumulating
dialogue history, in which earlier and updated constraints can remain
simultaneously accessible. Reliable interaction therefore requires resolving
which available constraints currently apply (\emph{select}) rather than
assuming that earlier information has been erased by an update
(\emph{append}). Whether the resolved information actually governs the
model's subsequent action, the \emph{deploy} step, is a separate behavioral
question.

\section{Earlier Dialogue Sources Remain Causally Relevant }
\label{sec:historical-causality}

Section~\ref{sec:state} showed that current and superseded values remain
accessible where they were introduced, but accessibility does not imply use.
This section asks whether these sources causally affect downstream behavior,
whether that influence changes after supersession, and how dependence on old
and new sources evolves after an update.

\subsection{Earlier Sources Carry Behaviorally Consequential Content}
\label{sec:causal-content}

The first intervention blocks later positions from attending to the source
span where a relevant value was introduced. Across eight models and
$1{,}186$--$1{,}195$ eligible instances per model, correct-value accuracy
falls from $.473$--$.605$ without intervention to $.202$--$.239$ under
source blocking, while length-matched same-turn, other-turn, and random-span
controls remain within $.007$ of the unmodified model. The disruption is concentrated in value-dependent behavior: frozen structural readouts agree with their unintervened predictions at $.981$--$.987$ for slot presence and above $.98$ for requests and domains, so blocking changes
value selection while leaving the tested structural signals largely intact.

To test whether source \emph{content} determines value preference, a source
containing $v_a$ is replaced with its counterpart from a matched counterfactual
context containing $v_b$. Across $678$--$839$ instances per model,
\[
    \log P(v_b)-\log P(v_a)
\]
shifts by $+6.55$ to $+14.22$ log-probability units and reverses sign in every model, while production of $v_b$ rises from $.085$--$.123$ to $.575$--$.670$. Random, unrelated-value, and wrong-span
substitutions do not reverse the preference. The earlier sources are
therefore not merely decodable: changing their accessibility or content can
redirect downstream value-dependent behavior.

\subsection{The Same Source Under Current and Superseded Status}
\label{sec:causal-validity}

Does the causal influence of the same historical source depend on its
current conversational validity? We construct $214$ matched
dialogue pairs per model. The target source occupies the same position and
contains the same value $v_1$ in both members; only the closing update cue
changes whether $v_1$ remains current or has been superseded by $v_2$.
Masking therefore removes identical source content from identical positions
under two different validity conditions.

Graded value support is
$z(H)=\log P(v_1\mid H)-\log P(v_2\mid H)$, with
$\Delta_{\mathrm{target}}(H)=z(H_{\mathrm{target\ mask}})-z(H)$, so
$\Delta_{\mathrm{target}}<0$ means that masking the source reduces relative
support for the value it contains.

\begin{figure}[!t]
    \centering
    \includegraphics[width=0.96\textwidth]{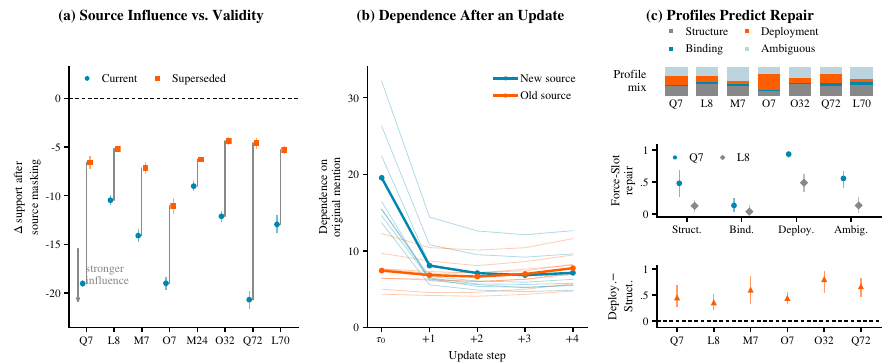}
    \caption{\textbf{Causal source influence and natural failure responses.}
    (a) Source influence changes with current conversational validity.
    (b) Dependence on earlier and updated sources shifts after an update but
    persists over subsequent turns.
    (c) Operational failure profiles predict different responses to the
    slot-forcing intervention.}
    \label{fig:causal-control}
\end{figure}

The same source has substantially greater influence while its value remains
current, but can remain influential after supersession. Averaged across
models, $\Delta_{\mathrm{target}}$ is $-14.66$ when $v_1$ remains current and
$-6.31$ after $v_1$ has been superseded. The graded effect is stronger in the
current condition in all eight models (Figure~\ref{fig:causal-control}a), while token-length-matched
random-span effects remain near zero.
The relevant evidence is the difference between validity conditions rather
than the sign of either masking effect alone: masking the only occurrence of
$v_1$ is expected to reduce its support in both twins. The comparison shows
that supersession attenuates this source dependence without necessarily
eliminating it.

To test whether this graded effect reaches the model's binary
current-value decision, let
$Y(H)=\mathbf{1}[\text{the model favors the currently valid value in }H]$,
and let $L_a^c$ denote the decrease in $Y$ after intervention $a$ under
condition $c\in\{\mathrm{cur},\mathrm{sup}\}$ (current, superseded). We
estimate the control-adjusted validity effect as
\begin{equation}
    D_{\mathrm{validity}}
    =
    \bigl(L_{\mathrm{target}}^{\mathrm{cur}}-L_{\mathrm{random}}^{\mathrm{cur}}\bigr)
    -
    \bigl(L_{\mathrm{target}}^{\mathrm{sup}}-L_{\mathrm{random}}^{\mathrm{sup}}\bigr),
\end{equation}
where the random intervention masks a token-length-matched non-value span.

The adjusted binary effect is positive on all eight models, ranging from
$+.051$ to $+.860$, with every dialogue-level bootstrap CI excluding zero (Table~\ref{tab:matched-validity-did}; descriptive macro mean
$+.582$). Thus, the same historical source affects both graded value support
and the model's current-value decision more strongly when its content remains
conversationally valid. This difference cannot be explained by the generic
effect of masking an equally long contextual span. Masking a different
slot's value is reported separately as an \emph{other-constraint}
intervention (Appendix~\ref{app:matched-validity}), not as a null control:
that span is itself part of the valid conversational state, and its removal
has measurable effects on several checkpoints. The matched comparison
establishes that conversational validity modulates source influence; it does
not by itself identify a distinct internal mechanism that performs validity
tracking.

\subsection{Dependence on Earlier Sources Changes After an Update}
\label{sec:causal-dynamics}

Finally, dependence on the original old- and new-value mentions is tracked
after an update. With $z(H)=\log P(v_{\mathrm{old}}\mid H)-\log P(v_{\mathrm{new}}\mid H)$,
let $C_{\mathrm{old}}(\tau)=z(H)-z(H_{\mathrm{mask\ old}})$,
$C_{\mathrm{new}}(\tau)=z(H_{\mathrm{mask\ new}})-z(H)$, and
$D(\tau)=C_{\mathrm{new}}(\tau)-C_{\mathrm{old}}(\tau)$, so that positive
$D$ indicates stronger dependence on the new source $\tau$ turns after the
update. At the update turn, $D(0)>0$ with confidence intervals excluding zero
in all eight models: dependence on the newly introduced source is initially
stronger (Figure~\ref{fig:causal-control}b). This advantage then shrinks. By
$\tau=+4$, the old--new difference is no longer statistically distinguishable
from zero in four models, while Llama-8B-IT,
Mistral-7B-IT, OLMo-7B-IT, and
OLMo-32B-IT show stronger dependence on the original old source.
The old-source effect itself remains substantial, ranging from $+4.73$ to
$+11.59$ log-probability units. These interventions concern the original mentions rather than every
downstream representation of their values, but they show that an update
does not sever the model from earlier information: a superseded mention
remains behaviorally consequential as the conversation continues.

\section{Natural Query Failures and Their Intervention Responses}
\label{sec:natural}
This section applies the measurements of Sections~\ref{sec:state}--\ref{sec:historical-causality}
to errors that arise naturally in unmodified closed-loop interaction,
asking whether query failures reflect weak structural support, incorrect value
resolution, failure to deploy otherwise-supported information, or interference
from superseded dialogue evidence. For each omitted constraint, we combine two measurements derived from the
preceding analyses. First, a calibrated linear readout at the decision site
estimates whether the missing slot is represented as active. Second, after
supplying the canonical slot key
domain: \{domain\}; \{slot\}:, we teacher-force candidate values and
measure the model's log-probability margin between the currently valid value
and its strongest genuinely different competitor. The latter is a behavioral
conditional-value score rather than a decision-site linear readout, and is
therefore distinct from the current-binding readout of
Section~\ref{sec:state-updates}.

We label an omission \textsc{Structure} when the slot readout is inactive,
\textsc{Binding} when the slot is supported but the conditional value margin
favors a competing value, and \textsc{Deployment} when the slot is supported
and the value margin does not favor a competing value, yet the constraint is
absent from the executable query.
Cases without a decisive assignment are labeled \textsc{Ambiguous}. These
profiles are operational diagnostics over the measured signals, not claims
about distinct internal modules or an exhaustive taxonomy of conversational
failure.

Natural omissions are heterogeneous across models. For
Qwen-7B-IT, $687$ omissions divide into $230$ structure, $38$
binding, $197$ deployment, and $222$ ambiguous cases;
Llama-8B-IT shows the same qualitative heterogeneity. Across the
replication panel, substantial mass appears in multiple profiles rather than
collapsing to a single failure type. The \textsc{Ambiguous} class remains
large, and \textsc{Binding} can contain relatively few examples on
individual models, so binding-specific rates are treated descriptively.

Do these diagnostic profiles predict different behavioral
responses? For omission errors, the \emph{slot-forcing intervention}
preserves the unmodified query prefix, appends only the missing slot key, and
lets the model freely generate its value. The intervention therefore supplies
the missing \emph{structure} without providing the correct value. Repair rates differ sharply by profile (Figure~\ref{fig:causal-control}c). On Qwen-7B-IT, slot forcing
repairs $.934$ of deployment-profile omissions, compared with $.478$ of
structure, $.132$ of binding, and $.554$ of ambiguous cases. On
Llama-8B-IT, the corresponding rates are $.486$, $.125$, $.036$,
and $.133$. The deployment-minus-structure contrast is positive with
confidence intervals excluding zero across the six models with
slot-forcing runs. This comparison is predictive rather than a causal identification of
separate mechanisms, since both the profile and the intervention condition on
the slot key; nevertheless, naturally occurring omissions are not
behaviorally interchangeable (controls in Appendix~\ref{app:natural-failures}).

Replaying the next query from closed-loop prefixes in which the unmodified
model emits a superseded value, with the superseded source masked, repairs
$17/28=.607$ of Qwen-7B-IT errors and $47/121=.388$ of
Llama-8B-IT errors, against $.07$--$.11$ under random or
unrelated-source masking; masking the currently valid source instead breaks
previously correct prefixes, while masking the superseded one rarely does
(Appendix~\ref{app:natural-failures}). This oracle-targeted diagnostic, not a
deployable repair, shows that superseded evidence causally drives a subset of
natural stale decisions. Relevant state can thus be weakly supported,
incorrectly resolved, or supported by the measured signals yet fail to reach
the executable action, motivating selective rather than wholesale correction
in Section~\ref{sec:control}.

\section{State--Action Control}
\label{sec:control}

\begin{figure}[!t]
\centering
\includegraphics[width=0.96\linewidth]
{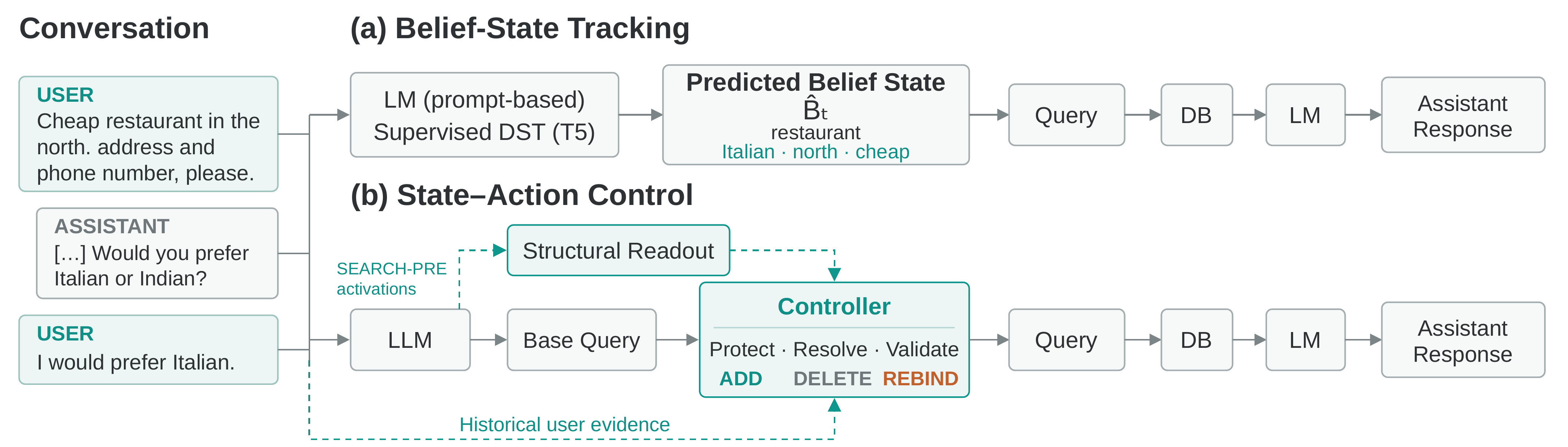}
\caption{\textbf{The State--Action Control interface.}
The Base LM's query is selectively edited using a structural readout and
visible historical evidence before database execution.}
\label{fig:control-interfaces}
\end{figure}

The \textbf{State--Action Controller} (\textsc{SAC}) starts from the Base LM's
own query, uses a supervised structural readout to propose targeted edits,
and resolves their values from visible user evidence; it neither modifies
activations nor constructs a complete belief state unlike other DST methods (Figure~\ref{fig:control-interfaces}). The design follows the diagnosis from earlier sections:
structure is readable at the decision site, binding is not, and values are
present in the history. The controller reads slot presence from the model and takes values from text.

\subsection{Controller and Operating Points}
\label{sec:control-method}

Let $q_{\mathrm{base}}$ be the Base LM's SEARCH query and $p_s$ the
Platt-calibrated posterior of a linear slot-presence readout for slot $s$,
fit as in Section~\ref{sec:state-structure} and read from the frozen decision
layer at the decision site. The controller
proposes additions and deletions according to

\begin{equation}
\begin{aligned}
\mathcal{A}_t
&= \{s : s \notin q_{\mathrm{base}},\; p_s > \tau_{\mathrm{add}}\},\\
\mathcal{D}_t
&= \{s : s \in q_{\mathrm{base}},\; p_s < \tau_{\mathrm{del}}\}.
\end{aligned}
\end{equation}
For $\tau_{\mathrm{del}}\leq p_s\leq\tau_{\mathrm{add}}$, the Base LM's
slot-presence decision is retained. Its domain is always preserved, and an
unparseable Base LM query is executed unchanged. Each proposal passes three safeguards. \emph{Evidence protection} vetoes
deletion when explicit user evidence continues to support the constraint.
\emph{Value resolution} binds an added slot to its latest explicit user-supported value and replaces an existing value only when newer explicit
evidence supplies a different value together with a fixed update cue.
\emph{Validation} rejects illegal edits and requires the compiled query to
reparse into the intended domain and fields; otherwise, the Base LM query is
retained.

Candidate evidence comes from visible user turns via frozen normalization,
alias, negation, and update-cue rules over a database-derived value
vocabulary; the Base LM supplies the domain. The controller sees no
simulator goal, reference state or query, system-side NLU, system turn, or
future turn. We evaluate three operating points
$(\tau_{\mathrm{add}},\tau_{\mathrm{del}})$: \emph{conservative} $(.90,.30)$
with a broad abstention region, \emph{intermediate} $(.70,.50)$, and
\emph{binary} $(.50,.50)$ with none. The intermediate setting gives the
highest mean exact-query accuracy and database equivalence on a 500-goal
development subset of Qwen-7B-IT and Llama-8B-IT, so we
fix $(.70,.50)$ before test evaluation
(Appendix~\ref{app:controller-selection}). Belief-state annotations
supervise only the structural readout; the controller carries no predicted
slot--value state across turns.

Offline replays on logged Base LM turns locate the source of the gain. SAC reaches exact-query accuracy $.708/.640$ on Qwen-7B-IT/Llama-8B-IT, versus at most $.586/.509$ for resolver-only, observable-text, shuffled-readout, and rate-matched random selectors (Appendix~\ref{app:selector-attribution}), while removing illegal slots from the Base LM queries reaches only $.445$ against SAC's $.621$ (Appendix~\ref{app:legality-filter}). The gain therefore requires dialogue-specific targeting by the learned readout. It also follows the profiles of Section~\ref{sec:natural}: SAC restores $.833$ of \textsc{Deployment} omissions but only $.131$ of \textsc{Binding} omissions (Appendix~\ref{app:profile-rescue}). Evidence protection blocks all proposed deletions of correct constraints (Appendix~\ref{app:d0-d1}).

\subsection{Closed-Loop Performance}
\label{sec:control-test}

Each configuration runs on 1,000 held-out goals per model. Training of the supervised methods is matched on the same eligible SEARCH decision sites, and all arms start from the same frozen goals and share the database, response model, simulator, NLU, and evaluator (Appendix~\ref{app:control-populations}); trajectories then evolve independently, since edited queries change database results and later turns. We report terminal task success (TS), paired by initial goal; exact-query accuracy, database equivalence, and slot F1 of the executed query, averaged over each arm's own SEARCH turns; preservation and repair, the fractions of already-correct and incorrect Base LM drafts that an arm leaves correct or corrects; destructive corrections per goal; and extra model generations per goal (Appendix~\ref{app:controller-diagnostics}).

Table~\ref{tab:control-main} compares SAC with two explicit-state baselines
and two references on the five-checkpoint macro; per-checkpoint results
and paired statistics appear in Appendix~\ref{app:d1-test-details}.  \textsc{Prompted DST} uses the same instruction-tuned LM to generate a
complete current slot--value state before deterministic query compilation.
Its prompt supplies the legal slot schema for the current domain; the
zero-shot variant uses no examples, and the $k{=}5$ and $k{=}10$ variants
prepend $k$ retrieved training-turn exemplars
(Appendix~\ref{app:prompted-dst}).  \textsc{T5-DST} uses a separately trained T5-small tracker for the same explicit-state interface (Appendix~\ref{app:t5-dst}). \textsc{Oracle Query} constructs the query from revealed current state while retaining the Base LM domain and
downstream policy, and is therefore a privileged reference rather than a
theoretical task-success ceiling.

\begin{table}[t]
\centering
\caption{\textbf{Held-out closed-loop MultiWOZ evaluation, five-checkpoint mean}
over Qwen-7B-IT, Llama-8B-IT, Mistral-7B-IT, OLMo-7B-IT, and OLMo-32B-IT,
$1{,}000$ goals per model and arm (per-checkpoint TS/Exact values in
Table~\ref{tab:controller-test-full}).
TS: terminal task success.
Exact, DB-eq: exact-query accuracy and database equivalence of the executed
query, averaged over each goal's SEARCH turns.
Slot F1: slot-level fidelity of the executed query against the reference state.
Preserv.: fraction of already-correct Base LM drafts that remain correct after
the arm acts.
Repair: fraction of incorrect Base LM drafts that are corrected.
Destr.: destructive corrections per goal.
Extra calls: model generations per goal beyond the Base LM's own query.
$\downarrow$: lower is better.}
\label{tab:control-main}
\setlength{\tabcolsep}{3.2pt}\small
\begin{tabular}{lrrrrrrrr}
\toprule
Method & TS & Exact & DB-eq & Slot F1 & Preserv. & Repair & Destr.\,$\downarrow$ & Extra calls\,$\downarrow$ \\
\midrule
Base LM                & .272 & .318 & .379 & .626 & 1.000 & .000 & .000 & 0 \\
Prompted DST (0-shot)  & .328 & .353 & .445 & .701 & .503 & .209 & .731 & 5.02 \\
Prompted DST ($k{=}5$) & .348 & .384 & .493 & .727 & .522 & .238 & .767 & 5.04 \\
Prompted DST ($k{=}10$)& .363 & .393 & .504 & .730 & .523 & .251 & .768 & 4.99 \\
T5-DST                 & \textbf{.372} & .549 & .575 & .798 & .654 & .402 & .650 & 4.98 \\
SAC-C                  & .349 & .534 & .569 & .784 & \textbf{.995} & .268 & \textbf{.010} & 0 \\
SAC-P                  & \textbf{.371} & \textbf{.621} & \textbf{.640} & .832 & .965 & .401 & .074 & 0 \\
SAC-B                  & .360 & .617 & .635 & \textbf{.836} & .918 & \textbf{.417} & .146 & 0 \\
\midrule
Oracle Query           & .423 & .983 & .907 & -- & -- & -- & -- & 0 \\
\bottomrule
\end{tabular}
\end{table}

The primary controller (SAC-P) improves on the Base LM on every checkpoint: TS from .272 to .371 (+.099, 95\% CI [+.089,+.111]), exact-query accuracy from .318 to .621, and database equivalence from .379 to .640. The three operating points trade preservation for repair (Table~\ref{tab:control-main}); the primary setting has the best tracking and TS and is used in all contrasts below. Against zero-shot Prompted DST, SAC-P gains +.043 TS, significant on three checkpoints, and +.267 exact-query accuracy. Retrieved exemplars raise prompted TS to .363 at k=10 but add only +.015 beyond k=5, and leave the fidelity gap intact (+.227 exact-query, positive on every checkpoint) and preservation near .52. Against the supervised T5-DST, SAC-P matches TS (.371 vs .372) and repair (.401 vs .402) and is higher on every other column: +.072 exact-query, +.065 database equivalence, slot F1 .832 vs .798, preservation .965 vs .654, and roughly nine times fewer destructive corrections (.074 vs .650 per goal), with no additional generation. The TS tie is heterogeneous: SAC-P leads on OLMo-32B-IT, T5-DST on OLMo-7B-IT, and three checkpoints are unresolved.

\paragraph{Interface and Inference Cost}
\textsc{SAC} adds no language-model generation, generated tokens, or additional prefill. Its structural readouts contain approximately
$35{,}800$--$51{,}200$ parameters per checkpoint, compared with $60.5$
million parameters for the shared fully fine-tuned T5 tracker. Prompted DST adds no separate parameters but requires $5.0$ additional Base-LM generations per goal at any $k$, with prefill rising from $3{,}100$ to
$6{,}100$ tokens per goal at $k{=}10$; T5-DST invokes its tracker $4.98$
times per goal (Appendix~\ref{app:d1-cost}). Measured on one A40 at batch size 1 over $300$ identical logged decisions,
\textsc{SAC}'s edit adds $5$--$7$\,ms and $8$\,MB per decision on
Qwen-7B-IT/Llama-8B-IT, against $655$--$1{,}202$\,ms for the Base LM's own draft,
$813$--$1{,}249$\,ms for one prompted-DST call at any $k$, and
$95$--$101$\,ms for a T5-DST call (Table~\ref{tab:controller-latency}).

We deliberately include no baseline that changes the deployed backbone's
weights, such as full SFT, LoRA adapters, RL fine-tuning against TS, or distillation. Even the lightest of these trains about $20$M adapter parameters
per 7B checkpoint ($0.3\%$ of the backbone, roughly $500\times$ the controller's
readout) through gradient passes over the backbone, yields a new checkpoint
per model, and either alters every query the model emits or adds a
generation per decision when used as a tracker. The research question here is how
reliably a fixed model's own representations and actions can
recover, so every comparator leaves the backbone unchanged or trains a
separate tracker on the same decision sites.

\paragraph{Robustness beyond template user language.}
Two experiments test whether SAC's gains depend on ConvLab's template user
NLG. First, we rerun the closed loop with every user utterance rewritten by
gpt-4o-mini under a faithfulness gate that preserves the user act's
values; goals, policy, database, and evaluator are unchanged
(Appendix~\ref{app:naturalized}). SAC's advantage over the Base LM is
unchanged: success $+.100$ ($[+.087,+.111]$), exact-query accuracy
$+.328$, and database equivalence $+.285$, against $+.099$, $+.303$, and
$+.260$ with template users; SAC's own scores shift by at most $.010$
(not significant on any checkpoint) and preservation stays at $.964$,
while the Base LM loses $.032$ exact-query accuracy. Second, we replay the
frozen controller on $4{,}490$ matched turns from $911$ human-authored
MultiWOZ~2.2 test dialogues, without execution, so this measures query-level
transfer rather than task success. SAC improves its Base LM by $+.158$
exact-query accuracy and $+.162$ database equivalence on the five-model
macro while breaking an initially correct query on $.0097$ of turns
(Appendix~\ref{app:human-transfer}).

\section{Discussion and Conclusion}
\label{sec:discussion}

Our results distinguish conversational information that remains accessible
from information that reliably governs action. Structural state is readable
near the decision site, exact values remain associated with their historical
sources, and matched interventions show that conversational validity
modulates those sources' causal influence. Natural failures accordingly
exhibit different operational profiles involving structure, binding, and
deployment. 

\paragraph{Positioning of State-Action Controller (SAC) }SAC tests whether these interpretability findings can support behavioral
control rather than proposing a state-of-the-art tracking architecture. A
small linear slot-presence readout selects targeted edits, visible dialogue
evidence resolves their values, and the Base LM's action is otherwise
preserved. Without complete state reconstruction, language-model
fine-tuning, or an additional generation call, SAC nearly doubles
exact-query accuracy and raises task success on every checkpoint.
T5-DST matches SAC's mean terminal success and repair, while SAC provides
higher query fidelity, better preservation of correct drafts, and far fewer
destructive corrections. SAC does so with orders of magnitude fewer trained
parameters than T5-DST and without the additional Base LM generations that
Prompted DST requires. Measured latent state can therefore improve action
without replacing the model's autoregressive generation.

These conclusions are limited to the readout interfaces, the interventions, operational diagnostics, and schema-aware interaction setting.
Reliable conversation nevertheless requires more than retaining state: a
system must resolve which information remains current and ensure that the
resulting state governs action.

\newpage
\subsubsection*{Reproducibility Statement}
Appendix~\ref{app:experimental-setup} documents the checkpoints and frozen
analysis layers (Table~\ref{tab:model-inventory}), dataset preprocessing and
evaluation populations, the closed-loop protocol and exact prompts
(Appendices~\ref{app:closed-loop-protocol} and~\ref{app:exact-prompts}),
readout fitting and calibration, every intervention definition, the exact
controller procedure (Appendix~\ref{app:d1-exact}), and the statistical
protocol with bootstrap seeds (Appendix~\ref{app:statistics}). Software
versions and hardware appear in Appendix~\ref{app:software}. The anonymous
supplement includes the model registry, readout-fitting and intervention
code, the controller implementation, frozen dataset and evaluation manifests
with hashes, and the evaluation scripts needed to reproduce the reported
analyses and results.

\subsubsection*{Ethics Statement}
This work studies conversational-state representations and behavioral
reliability using existing task-oriented dialogue benchmarks and simulated
interaction. We conduct no new human-subject data collection. Our
State--Action Controller is evaluated only in the schema-aware task-oriented
dialogue settings described in the paper and is not intended as a safety
mechanism for high-stakes deployment. As with the underlying language models
and datasets, the measurements and controller may inherit biases or
limitations present in their training data and benchmark domains. We report
these scope limitations explicitly and release the experimental artifacts
needed to audit the reported results.

\subsubsection*{AI Use Statement}
We used generative AI tools to aid and polish writing. AI assistance was used to
revise drafts of this paper for clarity, concision, and copy-editing. The
authors developed the research questions, experimental designs, analyses, and
interpretations, and reviewed every AI-assisted revision. We take
responsibility for the final content of this work, including all text, claims,
and artifacts. Additionally, generative AI (gpt-4o-mini) was used to rewrite user utterances for the naturalized-language evaluation described in Appendix~\ref{app:naturalized}.

\bibliography{references}
\bibliographystyle{iclr2027_conference}

\newpage
\appendix
\section{Experimental Setup and Reproducibility}
\label{app:experimental-setup}

This appendix documents the common experimental infrastructure used
throughout the paper. Experiment-specific populations, interventions,
controls, and full results are provided in the corresponding appendices.

\subsection{Models}
\label{app:models}

We evaluate eight instruction-tuned models spanning four model families and two
approximate scales per family. Table~\ref{tab:model-inventory} reports the
exact checkpoints and frozen analysis layers. All models are loaded in
\texttt{bfloat16} using the Hugging Face \texttt{transformers} library with
scaled-dot-product attention (\texttt{sdpa}). Models up to 9B parameters run
on a single GPU, 24--32B models are pipeline-sharded across two GPUs, and
70--72B models across four GPUs using \texttt{accelerate}.

\begin{table}[h]
\centering
\caption{\textbf{Instruction-tuned models used in the paper.}
Families: Qwen2.5~\citep{qwen2025qwen25technicalreport}, Llama~3.1~\citep{grattafiori2024llama3herdmodels}, Mistral~\citep{jiang2023mistral7b}, and OLMo~2~\citep{olmo20252olmo2furious}.
Model gives the name used in the text and Abbrev.\ the label used in the figures. The final column gives the
frozen model-specific layer used as the primary analysis layer. Exact checkpoint revisions corresponding to the executed runs
are recorded in the reproducibility package.}
\label{tab:model-inventory}
\scriptsize
\setlength{\tabcolsep}{3.2pt}
\begin{tabular}{l l l r r}
\toprule
Model & Abbrev. & Hugging Face checkpoint & Layers & Analysis layer \\
\midrule
Qwen-7B-IT
& Q7 & \texttt{Qwen/Qwen2.5-7B-Instruct}
& 28 & 20 \\

Qwen-72B-IT
& Q72 & \texttt{Qwen/Qwen2.5-72B-Instruct}
& 80 & 57 \\

Llama-8B-IT
& L8 & \texttt{meta-llama/Llama-3.1-8B-Instruct}
& 32 & 23 \\

Llama-70B-IT
& L70 & \texttt{meta-llama/Llama-3.1-70B-Instruct}
& 80 & 57 \\

Mistral-7B-IT
& M7 & \texttt{mistralai/Mistral-7B-Instruct-v0.3}
& 32 & 23 \\

Mistral-24B-IT
& M24 & \texttt{mistralai/Mistral-Small-24B-Instruct-2501}
& 40 & 28 \\

OLMo-7B-IT
& O7 & \texttt{allenai/OLMo-2-1124-7B-Instruct}
& 32 & 23 \\

OLMo-32B-IT
& O32 & \texttt{allenai/OLMo-2-0325-32B-Instruct}
& 64 & 45 \\
\bottomrule
\end{tabular}
\end{table}

For depth-dependent analyses, the frozen depth grid is
\[
\{0,\ .18,\ .36,\ .54,\ .71,\ .89,\ 1.0\}
\]
of model depth. The primary analysis layers in
Table~\ref{tab:model-inventory} were fixed before the downstream analyses. No
analysis layer, readout rank, or regularization strength was selected on any
test population: layers follow this frozen grid, and readout hyperparameters
are chosen on development splits (Appendix~\ref{app:readouts}).

\subsection{Datasets and Evaluation Populations}
\label{app:data}

\paragraph{MultiWOZ.}
Representation analyses use MultiWOZ~2.2~\citep{zang-etal-2020-multiwoz} with the EmoWOZ overlay~\citep{feng-etal-2022-emowoz},
materialized as one record per user turn followed by a system turn. The
original dialogue-level partition is preserved, yielding
8,432/999/999 dialogues and 56,728/7,371/7,363 user-turn examples in
train/development/test, respectively. The splits are dialogue-disjoint.
We use 35 belief-state slots across seven domains and 46 request keys. Slot
names, domains, and values are canonicalized using a frozen normalization map
before scoring.

Closed-loop experiments focus on the \texttt{hotel}, \texttt{restaurant},
and \texttt{attraction} domains. The environment uses the ConvLab-3~\citep{zhu-etal-2023-convlab} agenda-based
MultiWOZ user policy, template user NLG, BERT-based system-side NLU, RuleDST
for reference-state tracking, the MultiWOZ database, and
\texttt{MultiWozEvaluator}. Each experimental arm interacts with the user
simulator independently. Once two arms produce different queries, their
database results, responses, subsequent user turns, and trajectories may
diverge. Task success is therefore paired by the initial goal, whereas
turn-level query metrics may have different denominators across arms.

For controller development, we construct a frozen pool of 2,000 simulator
goals. A 1,000-goal development set and a disjoint 1,000-goal test set are
fixed before final evaluation. A nested 500-goal development subset is used
for controller selection and component ablations. The primary controller
operating point is selected on development data and then frozen before the
1,000-goal test evaluation.

\paragraph{Schema-Guided Dialogue.}
We additionally use Schema-Guided Dialogue (SGD)~\citep{rastogi2020scalablemultidomainconversationalagents} to test whether the
representation findings extend beyond MultiWOZ. Frozen preprocessing contains
345,464 turn records from 22,803 dialogues, spanning 45 services, 46 intents,
248 slot keys, and 156 request keys.

For the slot-update analysis, we identify turns at which the annotated value
of a slot changes from an earlier value to an updated value and both values
can be anchored to explicit dialogue mentions. The frozen population contains
812 update events from 686 dialogues, covering 33 services and 74 slots.
After requiring successful representation extraction at the relevant
positions, 762 events remain. The primary forced-choice localization
experiment fits on 458 events from 387 dialogues and evaluates on 304 events
from 258 disjoint dialogues.

\subsection{Controller Training and Test Population Reproducibility}
\label{app:control-populations}

The State--Action Controller is evaluated in the single-executable-domain setting. This isolates the state-resolution and deployment problem
studied by the controller from the separate problem of selecting among
simultaneously active executable domains.

For training, we construct a frozen manifest of MultiWOZ 2.2 SEARCH decision
sites whose reference state contains constraints for one executable domain. The training manifest contains 22,083 decision sites drawn from
5,932 of the 8,432 training dialogues; the development manifest contains
2,441 decision sites. The calibrated SAC readout is fit on every eligible
training item. Matched supervised baselines are constructed from the same (dialogue, turn) manifest.

The closed-loop evaluation uses the same scope. The frozen simulator-goal
universe contains 2,000 goals, all of which specify exactly one domain.
DEV500 and TEST1000 are disjoint 500 and 1,000-goal manifests drawn from this
universe, with the same domain mix. No selection used TEST1000, and every arm
is evaluated on the same goals. Across the 5,166 reference SEARCH decisions generated for TEST1000, none contains constraints from multiple domains.
Thus the controller's training population and deployed closed-loop evaluation share the same single-executable-domain interface.

\subsection{MultiWOZ Closed-Loop Protocol}
\label{app:closed-loop-protocol}

At each system decision point, the Base LM receives the last 64 rendered
dialogue lines followed by the frozen SEARCH instruction
(Appendix~\ref{app:exact-prompts}), shared by all experiments in
Sections~\ref{sec:natural}--\ref{sec:control}, and greedily emits one query
of semicolon-separated \texttt{slot: value} pairs beginning with the domain,
with 40 new tokens and at most 30 interaction turns.

The emitted query is parsed using the frozen query parser, restricted to legal
constrained slots for the parsed domain, and executed against the MultiWOZ
database. At most three matching entities are rendered back to the Base LM.
The model then generates the system response with a budget of 224 tokens.
All experimental arms share the same post-query database,
response-generation, NLU, simulator, and evaluation pipeline.

Task success is the terminal success signal returned by
\texttt{MultiWozEvaluator}. We additionally report two per-decision
diagnostics. \emph{Exact-query accuracy} requires the executed query to match
the reference query on the goal domain. \emph{Database equivalence} requires
the executed and reference queries to retrieve the same set of entity names.

\subsection{Prompted Belief-State Reconstruction}
\label{app:prompted-dst}

Prompted DST uses the same instruction-tuned LM as the Base LM and adds one
greedy state-generation call at each SEARCH decision. Given the same dialogue
history, the model is asked to emit the user's complete current slot--value
state as a JSON object. The prompt explicitly supplies the legal slot keys for
the current domain, so this baseline is schema-aware at inference time, but it
is not fitted on MultiWOZ belief-state annotations.

Parsed slot--value pairs are canonicalized and validated against the same
ontology and database-legality rules used elsewhere in the closed-loop
evaluation. The validated state is then deterministically compiled into the
same executable-query format and passed through the same database,
response-generation, and evaluation pipeline. The exact prompt is:

\begin{quote}
\small
\texttt{[STATE] Write the user's current state for the domain as one JSON object.}\\
\texttt{Allowed keys: \{slots\}. Include key/slots and values.}\\
\texttt{Output only the JSON object.}
\end{quote}

For example, in the restaurant domain, the allowed-key field is instantiated
as \texttt{food, pricerange, area, name}. Decoding is greedy and uses the
configuration reported in Appendix~\ref{app:exact-prompts}.

\paragraph{Few-shot variants.}
For $k\in\{5,10\}$ we prepend $k$ retrieved exemplars to the zero-shot prompt
above; $k{=}0$ reproduces the zero-shot arm exactly. The exemplar pool is the
MultiWOZ~2.2 training split ($56{,}728$ user turns from $8{,}432$ dialogues),
restricted to the $33{,}692$ turns whose executable state covers exactly one
domain (restaurant $13{,}702$, hotel $12{,}682$, attraction $7{,}308$; $7{,}238$
dialogues). Retrieval uses \texttt{all-mpnet-base-v2} sentence embeddings~\citep{reimers-gurevych-2019-sentence} with
cosine similarity over the key ``\texttt{[SYS]} \emph{last system utterance}
\texttt{[USER]} \emph{last user utterance}'', the retrieval-key format of
\citet{hu-etal-2022-context}, with no context state, so the arm remains
memoryless per turn. Candidates are restricted to the domain parsed from the
Base LM draft, so every exemplar's state uses only the allowed keys of the
current prompt. Each exemplar is rendered as \texttt{Example $n$}, its system
and user utterances, and its complete state as a JSON object over the domain's
constrained slots. The state target, compiler, legality rules, greedy decoding,
96-token budget, and 64-line history window are identical to the zero-shot
arm. Unlike \citet{hu-etal-2022-context}, the target is the complete current
state rather than a state change, and no previous state is conditioned on.

\subsection{Supervised T5-DST}
\label{app:t5-dst}

As a supervision-matched explicit-state baseline, we fully fine-tune
T5-small~\citep{3455716.3455856} to generate the complete current MultiWOZ slot--value state. The
tracker contains $60{,}506{,}624$ trainable parameters and is shared across
all five Base LM checkpoints of the closed-loop panel.

Training uses exactly the same $22{,}083$ eligible
$(\text{dialogue},\text{turn})$ decision sites used to fit the SAC structural
readout, drawn from the same $5{,}932$ training dialogues. The mapping between
the SAC and T5 manifests is complete, with no post-match filtering. The two
methods therefore receive the same training decisions but use different
objectives: SAC predicts slot presence, whereas T5-DST generates complete
slot--value state.

We train for five epochs with batch size 32, learning rate
$3\times10^{-4}$, AdamW with weight decay $.01$, maximum source length 512,
maximum target length 64, and seed 42. Selection uses $5{,}605$ eligible
MultiWOZ development examples and does not consult closed-loop outcomes.
The fifth epoch is selected by validation joint-goal accuracy
($.5260$; slot F1 $.8445$; exact-parse rate $.9938$).

At inference, T5-DST is invoked once per SEARCH decision, and its predicted
state is deterministically compiled into the same executable-query format
used by the other arms.

\subsection{Representation Readouts}
\label{app:readouts}

\paragraph{Structural accessibility.}
At each model's frozen analysis layer, we fit linear readouts for active
slots, domains, and requested information. Full-dimensional
readouts provide the primary accessibility measurements in the main text.
To test whether the structural signal contains a compact component, we also
evaluate supervised rank-32 projections against rank-matched PCA,
shuffled-supervision, and random-subspace controls. Fitting and evaluation
use dialogue-disjoint splits.

\paragraph{Calibrated slot-presence readout.}
The natural-failure diagnostics and State--Action Controller use a
model-specific linear slot-presence readout at the frozen decision layer.
For each model, this readout is fit on 22,083 training items and 2,441
development items using LBFGS with learning rate 1, at most 200 iterations,
and a strong-Wolfe line search. Probabilities are Platt-calibrated using
1,990 development turns comprising 9,373 slot instances.

For the natural-failure profiles only, a slot is treated as structurally
\textsc{active} when its calibrated posterior is at least $.70$,
\textsc{inactive} when it is at most $.30$, and \textsc{ambiguous}
otherwise. These diagnostic thresholds define the failure profiles in
Section~\ref{sec:natural}; they are distinct from the add/delete threshold
pairs used by the State--Action Controller.

\paragraph{Exact values.}
For each slot, we fit a multinomial linear probe over that slot's value
vocabulary. Probes are trained on MultiWOZ training dialogues and evaluated
on disjoint dialogues. Representations are read both at the earlier mention
where the value entered the conversation and at the later decision site.
Eligibility for the localization experiment depends on mention distance, not
on whether the value remains current, so the population contains both current
and subsequently superseded values.

\paragraph{Current binding.}
These are controlled synthetic dialogues instantiated with MultiWOZ slots and
values. The current-binding experiment uses matched counterfactual dialogues
containing the same candidate values but different update structure. Readout
selection is completed on development data before test inference. The primary
test condition contains 960 items, equally divided between
recency-congruent and recency-incongruent cases; the \texttt{correction} cue
family is withheld from fitting. Candidate value pairs and test cue
realizations are unseen during selection.

We additionally evaluate the unmodified model with a paired
\emph{counterfactual-pair accuracy} criterion: both members of a
counterfactual pair must select their respective currently valid values.
This pair-level measure is reported separately from ordinary item-level
current-value accuracy.

\subsection{Statistical Analysis}
\label{app:statistics}

Unless otherwise noted, uncertainty intervals use 2,000 bootstrap resamples
with seed 954100. Resampling follows the natural statistical unit of each
experiment: dialogues for representation and broad source-intervention
analyses, counterfactual pairs for matched-validity experiments, events
clustered by dialogue for temporal analyses, and simulator goals for
closed-loop MultiWOZ evaluation.

Closed-loop task success is a per-goal binary defined on the same initial
goal in every arm, so success comparisons are paired by initial simulator goal
and exact two-sided McNemar tests are computed on the discordant goals. Where
a comparison is repeated across the five-checkpoint panel, Holm correction is
applied across checkpoints \emph{within} each comparator family; the families
are never pooled. Current-binding item-level accuracies use Wilson intervals,
while paired counterfactual analyses use pair-level resampling.

Exact-query accuracy and database equivalence are computed per goal as the
mean over that goal's own SEARCH turns. Because edited queries change
database results and therefore subsequent turns, the arms' trajectories
diverge and their SEARCH turns are not matched. These quantities are paired
only by the shared initial goal and are reported as differences between
complete closed-loop interfaces rather than matched-turn effects; no McNemar
test is applied to them. The selector-attribution and profile-rescue analyses
(Appendices~\ref{app:selector-attribution} and~\ref{app:profile-rescue}) are the exceptions: they are offline
fixed-prefix evaluations over logged Base LM SEARCH turns, so every arm is
scored on exactly the same turns.

Macro quantities over the five-checkpoint panel are fixed-panel contrasts.
Each bootstrap draw resamples goals independently within each checkpoint,
computes that checkpoint's arm difference, and averages the five differences
with equal weight, so the resulting intervals describe uncertainty over goals
within this fixed panel rather than over a population of checkpoints.

\subsection{Model and Experiment Coverage}
\label{app:coverage}

Representation and causal analyses use all eight checkpoints. Later
behavioral experiments use the checkpoints with complete required runs for
that experiment; Table~\ref{tab:app-coverage} records coverage explicitly,
and no missing arm is imputed or included in the five-model macro.

\begin{table}[t]
\centering
\caption{\textbf{Model coverage across experimental components.}
Columns: representation readouts, causal source analyses, natural-failure profiles, slot forcing, and the complete closed-loop comparison, which requires the Base LM, Prompted DST,
T5-DST, and all three State--Action Controller operating points. ``Partial''
indicates that some of these arms are unavailable; no missing arm is imputed
or included in the five-model macro averages.}
\label{tab:app-coverage}

\scriptsize\setlength{\tabcolsep}{2pt}
\begin{tabular}{lccccc}
\toprule
Model &
Readouts & Causal & Profiles & Forcing & Closed loop \\
\midrule
Qwen-7B-IT
& Yes & Yes & Yes & Yes & Yes \\

Llama-8B-IT
& Yes & Yes & Yes & Yes & Yes \\

Mistral-7B-IT
& Yes & Yes & Yes & Yes & Yes \\

OLMo-7B-IT
& Yes & Yes & Yes & Yes & Yes \\

OLMo-32B-IT
& Yes & Yes & Yes & Yes & Yes \\

Mistral-24B-IT
& Yes & Yes & No & No & No \\

Qwen-72B-IT
& Yes & Yes & Yes & Yes & Partial \\

Llama-70B-IT
& Yes & Yes & Yes & No & Partial \\
\bottomrule
\end{tabular}
\end{table}

\subsection{Software, Hardware, and Reproducibility Artifacts}
\label{app:software}

Experiments use Python~3.10.19, PyTorch~2.5.1 with CUDA~12.1,
\texttt{transformers}~4.54.1, \texttt{accelerate}~0.34.2,
NumPy~1.26.4, SciPy~1.15.3, and scikit-learn~1.7.2. Attention uses
PyTorch scaled-dot-product attention; FlashAttention is not installed in the
audited environment.

Experiments were executed on NVIDIA A40 GPUs (46\,GB) and NVIDIA H200 GPUs
(144\,GB), without tensor parallelism.

\section{Full Representation and Accessibility Results}
\label{app:representation}

This appendix expands the accessibility analyses in Section~\ref{sec:state}.
The representation panel contains all eight instruction-tuned checkpoints,
independently of the five-checkpoint closed-loop comparison. Unless otherwise
specified, MultiWOZ readouts are fit on 56,728 user-turn examples from 8,432
training dialogues and evaluated on 7,371 examples from 999 disjoint
development dialogues. The model-specific analysis layers are listed in
Table~\ref{tab:model-inventory}; fitting procedures are described in
Appendix~\ref{app:readouts}. The counterfactual current-binding experiment
uses its own development-selected readouts and held-out test population.

\subsection{Structural State Readouts Across Eight Checkpoints}
\label{app:structural-readouts}

The structural signal at the decision site contains a compact component. At
rank $32$, supervised projections outperform rank-matched PCA and
shuffled-supervision controls in every evaluated model--task combination
(Figure~\ref{fig:accessibility}a). Slot-state projections retain
$51$--$58\%$ of full-dimensional F1 and request-state projections retain
$33$--$59\%$. Structural state is therefore not only accessible at the
decision site, but partly recoverable through a low-rank supervised
interface.

Table~\ref{tab:structural-accessibility} compares supervised rank-32
readouts with rank-matched PCA, shuffled-supervision, and random-subspace
controls, alongside full-dimensional readouts. All three structural tasks
use micro-F1. Random-subspace results summarize eight independently sampled
rank-32 subspaces.

\begin{table}[t]
\centering
\caption{\textbf{Structural state accessibility across eight checkpoints.}
All entries are micro-F1 on the MultiWOZ development split at the fixed
analysis layer. Supervised, PCA, and shuffled-supervision readouts use rank
32; Full uses the full-dimensional representation. Random reports the mean
over eight rank-32 random subspaces; ranges are included where available.}
\label{tab:structural-accessibility}
\scriptsize
\setlength{\tabcolsep}{3.2pt}
\textbf{(a) Slot presence}\par
\begin{tabular}{lrrrrr}
\toprule
Model & Supervised & PCA & Shuffled & Full & Random mean [min,max] \\
\midrule
Qwen-7B-IT     & .425 & .303 & .265 & .794 & .125 [.095,.161] \\
Llama-8B-IT    & .449 & .304 & .286 & .775 & .151 [.140,.167] \\
Mistral-7B-IT  & .408 & .247 & .229 & .802 & .117 [.108,.129] \\
OLMo-7B-IT     & .445 & .300 & .283 & .808 & .163 [.141,.197] \\
Mistral-24B-IT & .394 & .251 & .220 & .752 & .113 [.101,.129] \\
OLMo-32B-IT    & .444 & .300 & .278 & .789 & .165 [.148,.180] \\
Qwen-72B-IT    & .450 & .248 & .219 & .798 & .072 [.059,.088] \\
Llama-70B-IT   & .430 & .324 & .301 & .757 & .153 [.140,.176] \\
\bottomrule
\end{tabular}\par
\textbf{(b) Requested information}\par
\begin{tabular}{lrrrrr}
\toprule
Model & Supervised & PCA & Shuffled & Full & Random mean [min,max] \\
\midrule
Qwen-7B-IT     & .336 & .087 & .064 & .707 & .036 [.019,.046] \\
Llama-8B-IT    & .220 & .032 & .025 & .662 & .017 [.001,.030] \\
Mistral-7B-IT  & .430 & .154 & .124 & .780 & .064 [.052,.086] \\
OLMo-7B-IT     & .387 & .188 & .164 & .744 & .083 [.064,.116] \\
Mistral-24B-IT & .409 & .093 & .088 & .792 & .075 [.052,.106] \\
OLMo-32B-IT    & .473 & .224 & .213 & .802 & .140 [.114,.168] \\
Qwen-72B-IT    & .379 & .093 & .061 & .777 & .035 [.022,.058] \\
Llama-70B-IT   & .366 & .098 & .099 & .768 & .049 [.036,.064] \\
\bottomrule
\end{tabular}\par
\textbf{(c) Active domain}\par
\begin{tabular}{lrrrrr}
\toprule
Model & Supervised & PCA & Shuffled & Full & Random mean \\
\midrule
Qwen-7B-IT     & .812 & .748 & .715 & .956 & .591 \\
Llama-8B-IT    & .826 & .762 & .740 & .964 & .595 \\
Mistral-7B-IT  & .796 & .700 & .690 & .965 & .555 \\
OLMo-7B-IT     & .825 & .766 & .746 & .970 & .614 \\
Mistral-24B-IT & .795 & .696 & .670 & .956 & .559 \\
OLMo-32B-IT    & .812 & .742 & .723 & .976 & .614 \\
Qwen-72B-IT    & .813 & .688 & .677 & .926 & .517 \\
Llama-70B-IT   & .828 & .796 & .761 & .970 & .603 \\
\bottomrule
\end{tabular}
\end{table}

Supervised rank-32 readouts exceed both controls and every member of the
random-subspace bank for slot and request prediction in each checkpoint;
random subspaces already retain substantial domain information.
Full-dimensional readouts remain stronger, so structural information has a
recoverable low-rank component without occupying a dedicated 32-dimensional
register.

\subsection{Exact Historical Values Are More Accessible at Their Mentions}
\label{app:value-localization}

For each slot, a multinomial linear probe predicts historical-value identity
from either the historical mention representation or the decision-site
representation. Both positions are scored on the \emph{same} instances:
records are joined by dialogue, slot, and canonical value, so every scored
item contributes one measurement at each position and the comparison is
within-instance rather than between populations. The frozen candidate pool
contains $4{,}887$ instances whose historical mention occurs exactly three
turns before the decision site. The paired held-out population scored here is
the $3{,}255$ non-default instances within that pool, drawn from $922$
dialogues. The pool and the scored population are identical across all eight
checkpoints. Probes are fit on dialogue-disjoint training dialogues and
frozen before evaluation.

Eligibility depends on the mention-to-decision distance, not on current
validity. Both current and superseded values are included: the target is the
identity of a historical value, not whether it should govern the current
action.

\begin{table}[t]
\centering
\caption{\textbf{Paired exact-value localization at frozen analysis layers.}
The same $3{,}255$ held-out instances are scored at both positions, joined by
dialogue, slot, and canonical value. \textsc{Mention} uses the mean-pooled
historical value span and \textsc{Last} its final token; \textsc{Decision
site} reads the representation at the decision position, which occurs exactly
three turns after the mention ($k=3$). Gap is
\textsc{Mention}$-$\textsc{Decision site}, with a dialogue-level bootstrap
($2{,}000$ resamples, $922$ dialogues).}
\label{tab:app-value-localization-paired}
\small
\setlength{\tabcolsep}{4pt}
\begin{tabular}{lrrrrrr}
\toprule
Model & Layer & Mention & Last & Decision site ($k=3$) & Gap & 95\% CI \\
\midrule
Qwen-7B-IT             & 20 & .9696 & .9241 & .1819 & $+.7877$ & $[+.773, +.802]$ \\
Llama-8B-IT           & 23 & .9788 & .9730 & .2399 & $+.7389$ & $[+.722, +.756]$ \\
Mistral-7B-IT        & 23 & .9920 & .9429 & .1837 & $+.8083$ & $[+.794, +.823]$ \\
OLMo-7B-IT         & 23 & .9871 & .9548 & .2147 & $+.7724$ & $[+.757, +.787]$ \\
Mistral-24B-IT & 28 & .9926 & .9806 & .1862 & $+.8065$ & $[+.792, +.821]$ \\
OLMo-32B-IT        & 45 & .9849 & .9662 & .2055 & $+.7794$ & $[+.765, +.794]$ \\
Qwen-72B-IT            & 57 & .9613 & .9561 & .2018 & $+.7594$ & $[+.745, +.775]$ \\
Llama-70B-IT          & 57 & .9853 & .9810 & .3127 & $+.6725$ & $[+.655, +.689]$ \\
\bottomrule
\end{tabular}
\end{table}

Historical-value identity is recovered with $.961$--$.993$ accuracy from the
mean-pooled mention span, compared with $.182$--$.313$ at the decision site
exactly three turns later. Every paired gap interval excludes zero. High
mention-site accuracy is expected in part because that representation is
computed where the value's surface form appears, and a causal transformer
cannot revise that position once later turns are appended. The result
establishes an accessibility asymmetry under the tested linear interfaces; it
does not imply that exact values are absent elsewhere in the model.

A layer-0 decision-site control is approximately $.022$ in the original
four-checkpoint analysis, confirming that the non-trivial decision-site
accuracies reflect contextual rather than lexical information at a fixed
template position. Categorical values are more recoverable at the decision
site than entity/location or time/numeric values, while all three strata are
near ceiling at their mentions. These additional controls have narrower model
coverage than Table~\ref{tab:app-value-localization-paired}.

\paragraph{Value readability at the emission position.}
On the same instances, we append the teacher-forced prefix
\texttt{domain: \{dom\}; \{slot\}:} to the decision-site prompt and fit the
identical probe at its final token, the position from which the value would
be emitted. Table~\ref{tab:emission-position} reports all / non-default
accuracy at the frozen layer; the decision-site column reproduces
Table~\ref{tab:app-value-localization-paired}. Categorical slots reach
$.82$--$.98$ at the emission position, while names, times, and booking counts
remain at $.03$--$.36$ (Qwen-7B-IT). The emission position is
teacher-forced, so this measures what is readable given the key, not whether
the model produces the key.

\begin{table}[t]
\centering\small
\caption{\textbf{Exact-value probe at the decision site and the emission position on identical instances.}
All / non-default accuracy on the $4{,}887$ / $3{,}255$ held-out instances
of Table~\ref{tab:app-value-localization-paired}, same probe, layer, and
training discipline. The last column is the paired non-default difference
with a dialogue-level bootstrap ($2{,}000$ resamples); every interval
excludes zero.}
\label{tab:emission-position}
\setlength{\tabcolsep}{4pt}
\begin{tabular}{lccc}
\toprule
Model & Decision site & Emission position & Emission $-$ decision (non-default) \\
\midrule
Qwen-7B-IT             & .295 / .182 & .708 / .618 & $+.436$ $[+.417,+.453]$ \\
Llama-8B-IT           & .363 / .240 & .926 / .908 & $+.668$ $[+.651,+.685]$ \\
Mistral-7B-IT        & .308 / .184 & .903 / .872 & $+.688$ $[+.672,+.705]$ \\
OLMo-7B-IT         & .334 / .215 & .900 / .873 & $+.658$ $[+.638,+.676]$ \\
Mistral-24B-IT & .299 / .186 & .901 / .872 & $+.686$ $[+.669,+.702]$ \\
OLMo-32B-IT        & .328 / .206 & .839 / .786 & $+.580$ $[+.562,+.597]$ \\
Qwen-72B-IT            & .321 / .202 & .811 / .744 & $+.542$ $[+.526,+.558]$ \\
Llama-70B-IT          & .426 / .313 & .955 / .942 & $+.629$ $[+.611,+.646]$ \\
\bottomrule
\end{tabular}
\end{table}

\subsection{Replication on Schema-Guided Dialogue}
\label{app:sgd-accessibility}

We examine value accessibility on SGD events in which a service--slot value
changes and both the old and new values can be anchored to explicit mentions.
Source representations are taken at the final token of each value span;
the decision-site representation is taken at the end of the visible context.

The initial population contains 812 events from 686 dialogues. Successful
site extraction leaves 762 events: 458 events from 387 dialogues for fitting
and 304 events from 258 disjoint dialogues for evaluation. Each forced-choice
task distinguishes the event's two candidate values, matched on service,
slot, and dialogue; binary chance accuracy is $.5$.

\begin{table}[t]
\centering
\caption{\textbf{Historical-value accessibility on SGD slot-update events.}
Forced-choice accuracies on 304 events from 258 held-out dialogues, with
95\% bootstrap intervals. Old and new source columns measure identification
of the value at the corresponding mention. Decision-site current binding is
the fraction of events on which the readout favors the updated value, not
its mean predicted probability.}
\label{tab:sgd-source-accessibility}
\scriptsize
\setlength{\tabcolsep}{3pt}
\begin{tabular}{lccc}
\toprule
Model & Old source & New source & Decision-site current binding \\
\midrule
Qwen-7B-IT     & .868 [.829,.906] & .888 [.853,.924] & .569 [.515,.625] \\
Llama-8B-IT    & .888 [.852,.923] & .918 [.886,.948] & .477 [.422,.537] \\
Mistral-7B-IT  & .885 [.848,.920] & .905 [.870,.938] & .507 [.448,.564] \\
OLMo-7B-IT     & .898 [.862,.933] & .911 [.878,.942] & .507 [.446,.563] \\
Mistral-24B-IT & .895 [.860,.927] & .941 [.913,.965] & .566 [.514,.622] \\
OLMo-32B-IT    & .888 [.854,.922] & .885 [.846,.919] & .480 [.423,.538] \\
Qwen-72B-IT    & .891 [.855,.925] & .914 [.881,.947] & .572 [.517,.626] \\
Llama-70B-IT   & .891 [.857,.924] & .914 [.883,.946] & .576 [.522,.629] \\
\bottomrule
\end{tabular}
\end{table}

Cross-site generalization is also high ($.786$--$.878$), consistent with
readouts recovering value content rather than only a source-position
signature. Because an earlier position cannot incorporate later tokens,
persistence there does not by itself establish an updating mechanism;
Section~\ref{sec:historical-causality} tests whether these sources still
influence decisions.

\subsection{Counterfactual Current-Binding Readouts}
\label{app:current-binding-readouts}

We next ask whether a fixed decision-site readout identifies the currently
valid value when the same candidates occur under different update
instructions. The primary test condition contains 960 items, equally divided
between recency-congruent and recency-incongruent cases. The correction cue
family is withheld from fitting and contributes 342 test items, including
171 recency-incongruent items. Candidate pairs and test cue realizations are
unseen during readout selection.

\begin{table}[t]
\centering
\caption{\textbf{Decision-site current-binding readout.}
Item-level accuracy with Wilson 95\% intervals. Analysis layer, rank, and
weight decay (WD) are fixed on development data before test evaluation.
The full test has 960 items, each recency stratum has 480, the held-out
correction family has 342, and its recency-incongruent subset has 171.}
\label{tab:current-binding-readout}

\scriptsize\setlength{\tabcolsep}{2pt}
\begin{tabular}{lrrrccccc}
\toprule
Model & L & Rank & WD &
Overall & Congruent & Incongruent & Correction & Corr.$\times$Incong. \\
\midrule
Qwen-7B-IT     & 20 & 8   & $10^{-4}$ & .510 [.479,.542] & .527 [.482,.571] & .494 [.449,.538] & .488 [.436,.541] & .474 [.400,.548] \\
Llama-8B-IT    & 23 & 16  & $10^{-4}$ & .557 [.526,.588] & .538 [.493,.582] & .577 [.532,.620] & .474 [.421,.527] & .456 [.383,.531] \\
Mistral-7B-IT  & 23 & 2   & $10^{-4}$ & .534 [.503,.566] & .429 [.386,.474] & .640 [.596,.681] & .503 [.450,.556] & .602 [.528,.673] \\
OLMo-7B-IT     & 23 & 16  & $10^{-4}$ & .532 [.501,.564] & .494 [.449,.538] & .571 [.526,.614] & .506 [.453,.558] & .503 [.429,.577] \\
Mistral-24B-IT & 28 & 512 & $10^{-3}$ & .517 [.485,.548] & .544 [.499,.588] & .490 [.445,.534] & .500 [.447,.553] & .515 [.440,.588] \\
OLMo-32B-IT    & 45 & 8   & $10^{-3}$ & .591 [.559,.621] & .556 [.512,.600] & .625 [.581,.667] & .515 [.462,.567] & .550 [.475,.622] \\
Qwen-72B-IT    & 57 & 64  & $10^{-3}$ & .540 [.508,.571] & .515 [.470,.559] & .565 [.520,.608] & .515 [.462,.567] & .538 [.463,.611] \\
Llama-70B-IT   & 57 & 16  & $10^{-4}$ & .537 [.505,.568] & .548 [.503,.592] & .525 [.480,.569] & .538 [.485,.590] & .521 [.446,.594] \\
\bottomrule
\end{tabular}
\end{table}

Overall accuracy ranges from $.510$ to $.591$, recency-incongruent accuracy
from $.490$ to $.640$, and held-out correction accuracy from $.474$ to
$.538$. Chance, first-position, bag-of-values, and overall-recency baselines
are $.5$; the recency heuristic scores zero on recency-incongruent items.
The readout therefore does not robustly recover current binding across these
controlled conditions. This conclusion concerns the tested linear interface,
not the presence or absence of current-binding information in the model.

\paragraph{Nonlinear readouts.}
To test whether the near-chance result reflects the linear form of the
readout, we fit two nonlinear readouts under the identical protocol: the same
DEV items, anchor layer, decision-site extraction, value embeddings, inner
split by value pair, and selection on dev-sel recency-incongruent accuracy,
followed by a single evaluation on the same 960 TEST items. \textsc{MLP-H}
replaces the bilinear map with a two-layer network on the state,
$s(h,v)=\mathrm{MLP}(h)^{\top}e_v$; \textsc{MLP-J} scores state and candidate
jointly, $s(h,v)=\mathrm{MLP}([h;e_v])$. Hidden width $\{64,256,1024\}$ and
weight decay $\{10^{-4},10^{-3},10^{-2}\}$ are selected on development data.
Neither improves materially on the linear readout: on every one of the four
evaluated checkpoints the nonlinear overall accuracies lie within $.04$ of the
linear one, and none exceeds $.56$ overall or $.58$ on recency-incongruent items
(Table~\ref{tab:mlp-readout}), while the unmodified models' item-level
accuracy on the same items is $.434$--$.580$. The tested decision-site
interfaces, linear or not, therefore do not expose current binding; the
models' own competence is not thereby denied.

\begin{table}[t]
\centering
\caption{\textbf{Linear versus nonlinear current-binding readouts on the same
960 test items.} Overall and recency-incongruent item-level accuracy; the
linear readout is the frozen one of Table~\ref{tab:current-binding-readout},
scored on identically extracted states. Width is the development-selected
hidden size.}
\label{tab:mlp-readout}
\scriptsize
\setlength{\tabcolsep}{3pt}
\begin{tabular}{lcccccccc}
\toprule
 & \multicolumn{2}{c}{Linear} & \multicolumn{3}{c}{MLP-H} & \multicolumn{3}{c}{MLP-J} \\
\cmidrule(lr){2-3}\cmidrule(lr){4-6}\cmidrule(lr){7-9}
Model & All & Incong. & All & Incong. & Width & All & Incong. & Width \\
\midrule
Qwen-7B-IT & .510 & .494 & .503 & .500 & 64 & .498 & .496 & 256 \\
Llama-8B-IT & .557 & .577 & .559 & .537 & 256 & .523 & .537 & 1024 \\
Mistral-7B-IT & .534 & .640 & .494 & .502 & 64 & .528 & .571 & 1024 \\
OLMo-7B-IT & .532 & .571 & .500 & .500 & 64 & .514 & .515 & 64 \\
\bottomrule
\end{tabular}
\end{table}

\subsection{Model Behavior on the Same Counterfactuals}
\label{app:readout-behavior-disagreement}

Pair accuracy requires both members of a twin to select their respective
currently valid values, which prevents an always-same-candidate strategy from
succeeding; it is stricter than item-level accuracy.
Table~\ref{tab:readout-behavior-joint} reports both, together with a
cross-tabulation of readout and model correctness on the same 960 items.
Qwen-72B-IT and Llama-70B-IT reach pair accuracies of $.598$ and $.721$, and
$.731$ and $.830$ on the held-out correction family. Pair accuracy is far stricter than item accuracy: Qwen-7B-IT has pair
accuracy $.027$ but item accuracy $.434$. Among readout errors, the model
still selects the correct current value in $.774$ of cases for
Mistral-24B-IT, $.758$ for Qwen-72B-IT, and $.820$ for Llama-70B-IT, so weak
recoverability through this fixed readout does not imply absent
current-binding competence. These measurements do not determine whether
additional information is nonlinear, distributed across positions, or
constructed during decoding.

\begin{table}[t]
\centering
\caption{\textbf{Model behavior and readout correctness on the same
counterfactual items.} Pair accuracy (480 pairs) requires both members to
select their current values; item accuracy is over the 960 items. RD denotes
the fixed decision-site readout and MB the unmodified model's current-value
behavior; counts cross-tabulate item-level correctness.}
\label{tab:readout-behavior-joint}
\scriptsize
\setlength{\tabcolsep}{1.5pt}
\begin{tabular}{lccrrrrc}
\toprule
Model & Pair acc. [95\% CI] & Item acc. &
RD$\checkmark$/MB$\checkmark$ &
RD$\checkmark$/MB$\times$ &
RD$\times$/MB$\checkmark$ &
RD$\times$/MB$\times$ &
$P(\mathrm{MB}\checkmark\mid\mathrm{RD}\times)$ \\
\midrule
Qwen-7B-IT     & .027 [.015,.042] & .434 & 227 & 263 & 190 & 280 & .404 \\
Llama-8B-IT    & .210 [.173,.248] & .463 & 277 & 258 & 167 & 258 & .393 \\
Mistral-7B-IT  & .258 [.219,.298] & .580 & 273 & 240 & 284 & 163 & .635 \\
OLMo-7B-IT     & .090 [.065,.115] & .467 & 243 & 268 & 205 & 244 & .457 \\
Mistral-24B-IT & .588 [.542,.629] & .784 & 394 & 102 & 359 & 105 & .774 \\
OLMo-32B-IT    & .410 [.367,.456] & .689 & 407 & 160 & 254 & 139 & .646 \\
Qwen-72B-IT    & .598 [.554,.640] & .756 & 391 & 127 & 335 & 107 & .758 \\
Llama-70B-IT   & .721 [.677,.760] & .842 & 443 & 72  & 365 & 80  & .820 \\
\bottomrule
\end{tabular}
\end{table}

\section{Causal Interventions on Earlier Dialogue Sources}
\label{app:historical-causality}
This appendix reports the complete intervention results of
Section~\ref{sec:historical-causality}.
\subsection{Blocking Historical Value Sources}
\label{app:source-blocking}
We first test whether a historical user value remains behaviorally
consequential at a later decision point. Eligible examples contain a
non-default value introduced at least three turns earlier, with no subsequent
system restatement of that value. The intervention blocks attention to the
tokens corresponding to the original historical value span.
Our primary \emph{broad} intervention prevents every later position from
attending to the selected source span, at every layer and head. Three
token-length-matched controls block a span from the same user turn, an earlier
different user turn, or a random contiguous location in the rendered prompt.
All interventions are evaluated on the same instances.
\begin{table}[t]
\centering
\caption{\textbf{Broad blocking of historical value sources.}
\textsc{Unmodified} is value-selection accuracy without intervention.
The final three columns report broad-block accuracy minus the corresponding
matched-control accuracy, with 95\% bootstrap intervals.}
\label{tab:source-blocking}

\scriptsize\setlength{\tabcolsep}{2pt}
\begin{tabular}{lrrrrrcc}
\toprule
Model & $n$ & Dial. & Unmodified & Blocked &
vs.\ same turn & vs.\ other turn & vs.\ random \\
\midrule
Qwen-7B-IT     & 1187 & 533 & .473 & .202 &
-.269 [-.298,-.239] & -.270 [-.299,-.241] & -.276 [-.306,-.247] \\
Llama-8B-IT    & 1194 & 534 & .569 & .225 &
-.343 [-.375,-.311] & -.346 [-.379,-.317] & -.343 [-.374,-.313] \\
Mistral-7B-IT  & 1186 & 532 & .498 & .206 &
-.292 [-.321,-.261] & -.293 [-.325,-.263] & -.295 [-.325,-.266] \\
OLMo-7B-IT     & 1192 & 533 & .534 & .205 &
-.323 [-.352,-.293] & -.326 [-.358,-.296] & -.325 [-.358,-.293] \\
Mistral-24B-IT & 1186 & 531 & .605 & .239 &
-.358 [-.388,-.327] & -.365 [-.394,-.334] & -.362 [-.392,-.330] \\
OLMo-32B-IT    & 1195 & 534 & .577 & .234 &
-.341 [-.374,-.308] & -.346 [-.379,-.315] & -.342 [-.374,-.312] \\
Qwen-72B-IT    & 1187 & 532 & .534 & .213 &
-.324 [-.355,-.292] & -.324 [-.356,-.295] & -.322 [-.353,-.291] \\
Llama-70B-IT   & 1195 & 534 & .572 & .232 &
-.338 [-.369,-.306] & -.341 [-.373,-.311] & -.338 [-.371,-.308] \\
\bottomrule
\end{tabular}
\end{table}
Broad source blocking lowers value-selection accuracy from
$.473$--$.605$ to $.202$--$.239$ across the eight checkpoints. Each of the
24 model-by-control contrasts excludes zero, whereas the matched controls
themselves change the unmodified baseline by at most $.0067$ in absolute
value. The matched controls support a source-specific effect relative to the tested
perturbations. Broad blocking can nevertheless affect downstream contextual
computations as well as direct retrieval, so it does not isolate a single
retrieval edge or constitute complete erasure of the value.
Structural readout predictions remain highly stable under the intervention:
slot-presence self-agreement is $.981$--$.987$, request self-agreement
$.9997$--$.9999$, and domain self-agreement $.991$--$.996$. These quantities
are defined relative to the model's own unmodified readout predictions and
should not be compared with the gold-label structural F1 values reported in
Appendix~\ref{app:representation}.
\subsection{Substituting Historical Value Content}
\label{app:source-substitution}
Blocking establishes that the historical source matters; substitution tests
whether the content carried by that source determines which value is selected.
We construct a counterfactual dialogue that differs only in the value stated
for the same slot. At a frozen early depth, the full residual hidden states
over the original historical value span are replaced by those from the
counterfactual value span. The donor value is required to have the same token
count and is selected by a seeded procedure fixed before inference.
The endpoint is
\[
m = \log P(v_b \mid H)-\log P(v_a \mid H).
\]
where $v_a$ is the originally stated value and $v_b$ the substituted value.
\begin{table}[t]
\centering
\caption{\textbf{Historical-value substitution.}
Margin is $\log P(v_b)-\log P(v_a)$. The final two columns report the
fraction of generations that express the substituted and original value,
respectively, before and after patching.}
\label{tab:source-substitution}

\scriptsize\setlength{\tabcolsep}{2pt}
\begin{tabular}{lrrrrccc}
\toprule
Model & $n$ & Dial. & Unmod. & Patched &
Shift [95\% CI] & Says $v_b$ & Says $v_a$ \\
\midrule
Qwen-7B-IT     & 798 & 424 & -4.170 & +4.384 & +8.55 [+7.95,+9.16] & .123$\rightarrow$.600 & .526$\rightarrow$.089 \\
Llama-8B-IT    & 836 & 443 & -4.914 & +3.798 & +8.71 [+8.13,+9.27] & .085$\rightarrow$.624 & .620$\rightarrow$.086 \\
Mistral-7B-IT  & 678 & 390 & -4.972 & +4.426 & +9.40 [+8.68,+10.11] & .118$\rightarrow$.575 & .544$\rightarrow$.102 \\
OLMo-7B-IT     & 838 & 444 & -4.943 & +4.088 & +9.03 [+8.44,+9.62] & .099$\rightarrow$.619 & .585$\rightarrow$.086 \\
Mistral-24B-IT & 685 & 389 & -3.465 & +3.083 & +6.55 [+6.04,+7.04] & .114$\rightarrow$.666 & .634$\rightarrow$.091 \\
OLMo-32B-IT    & 839 & 444 & -5.962 & +4.723 & +10.69 [+9.99,+11.40] & .094$\rightarrow$.614 & .616$\rightarrow$.113 \\
Qwen-72B-IT    & 801 & 425 & -7.387 & +6.832 & +14.22 [+13.37,+15.11] & .097$\rightarrow$.670 & .653$\rightarrow$.105 \\
Llama-70B-IT   & 839 & 444 & -4.927 & +4.339 & +9.27 [+8.71,+9.81] & .094$\rightarrow$.646 & .615$\rightarrow$.066 \\
\bottomrule
\end{tabular}
\end{table}
The intervention reverses the value margin in every checkpoint. The mean
shift ranges from $+6.55$ to $+14.22$, while production of the substituted
value rises from $.085$--$.123$ to $.575$--$.670$.
Three controls separate value substitution from generic hidden-state
perturbation: a norm-matched random vector at the true span, an unrelated
active-slot representation at the true span, and the correct counterfactual
content inserted at a neighboring non-value span.
Patch minus control margins range from $+4.01$ to $+10.75$, and all 24 intervals exclude zero. The control margins remain negative, although they are less negative than
the unmodified margins. Thus, these perturbations weaken preference for the
original value without reproducing the sign reversal induced by substitution
at the true source. This supports content- and location-sensitive influence,
without implying that the intervention changes only the target value.
The same intervention leaves the frozen structural readout nearly unchanged:
presence agreement is $.988$--$.993$, request agreement is $.9998$--$.9999$,
and domain agreement is $.996$--$.999$. Agreement for other active-slot values is lower ($.908$--$.944$), indicating
that the intervention can also have collateral effects.
\subsection{Matched Current-versus-Superseded Sources}
\label{app:matched-validity}
The preceding experiments show that historical sources are causally relevant,
but do not distinguish currently valid from superseded evidence. We therefore
construct 214 counterfactual twin pairs per checkpoint. Each pair contains
the same two value mentions at the same positions; only the final update cue
changes whether the earlier source remains current or has been superseded.
No explicit belief state is shown. Let $v_1$ denote the value at the
fixed target source and $v_2$ the alternative. In both twins we use
\[
    z(H)=\log P(v_1\mid H)-\log P(v_2\mid H),
    \qquad
    \Delta_{\mathrm{mask}}(H)=z(H_{\mathrm{mask}})-z(H).
\]
Thus, $\Delta_{\mathrm{mask}}<0$ means that masking the source reduces support
for the value it carries, irrespective of whether that value remains valid.
\begin{table}[t]
\centering
\caption{\textbf{Matched current-versus-superseded source influence.}
The same historical source is current in one twin and superseded in the other.
The paired difference is Superseded$-$Current, so positive values indicate
stronger causal influence while the source is current.}
\label{tab:matched-source-validity}
\scriptsize
\setlength{\tabcolsep}{3pt}
\begin{tabular}{lccc}
\toprule
Model &
Current $\Delta_{\mathrm{mask}}$ &
Superseded $\Delta_{\mathrm{mask}}$ &
Superseded$-$Current \\
\midrule
Qwen-7B-IT
& -19.009 [-19.34,-18.67]
& -6.591 [-7.16,-6.01]
& +12.418 [+11.82,+12.99] \\
Llama-8B-IT
& -10.458 [-10.91,-9.98]
& -5.163 [-5.40,-4.92]
& +5.295 [+4.87,+5.70] \\
Mistral-7B-IT
& -14.092 [-14.72,-13.47]
& -7.154 [-7.75,-6.57]
& +6.938 [+6.22,+7.72] \\
OLMo-7B-IT
& -18.982 [-19.58,-18.38]
& -11.034 [-11.79,-10.29]
& +7.948 [+7.30,+8.64] \\
Mistral-24B-IT
& -9.015 [-9.48,-8.57]
& -6.275 [-6.50,-6.05]
& +2.739 [+2.33,+3.16] \\
OLMo-32B-IT
& -12.111 [-12.64,-11.58]
& -4.371 [-4.74,-4.06]
& +7.741 [+7.19,+8.32] \\
Qwen-72B-IT
& -20.667 [-21.52,-19.80]
& -4.574 [-5.10,-4.09]
& +16.092 [+15.15,+16.96] \\
Llama-70B-IT
& -12.941 [-13.77,-12.05]
& -5.284 [-5.61,-4.97]
& +7.657 [+6.72,+8.57] \\
\bottomrule
\end{tabular}
\end{table}
Masking the target source reduces support for its value in both contexts for
all eight checkpoints. Mean $\Delta_{\mathrm{mask}}$ is $-14.659$ while the
source is current and $-6.306$ after it is superseded. The paired difference
excludes zero in all eight checkpoints, and every one shows weaker influence
after supersession.
Crucially, superseded sources are not rendered inert: their masking effect is
negative in every checkpoint. Under the tested intervention, this is inconsistent with complete exclusion
of the superseded source from downstream value selection; it does not exclude
other representations of the current state.
The control results also limit the stronger interpretation that this
difference identifies a dedicated validity mechanism.
Across the eight checkpoints, the Superseded$-$Current difference is $-.091$ to $+.046$ under the random mask, $+.134$ to $+10.113$ under other-constraint masking, and $+17.653$ to $+60.416$ under current-source masking. The random-span interaction is effectively null. The other-constraint
interaction is not, and should not be read as one: that span carries a
different slot's value, which is itself part of the valid conversational
state, so removing it deletes real task content rather than neutral filler.
We therefore use the token-length-matched random non-value span as the
nonspecific control throughout, and report other-constraint masking only as a
secondary intervention. The current-source mask is not an independent
replication in the current twin because it coincides with the target-source
mask there by construction.

\paragraph{Control-adjusted current-value decision.}
The graded analysis above measures relative support. We additionally test
whether validity status reaches the model's binary current-value decision.
Let $Y(H)$ indicate that the model favors the currently valid value, and let
$L_a^c$ denote the decrease in $Y$ after intervention $a$ under condition
$c\in\{\mathrm{current},\mathrm{superseded}\}$. The control-adjusted validity
effect is the difference-in-differences
\[
    D_{\mathrm{validity}}
    =
    \bigl(L_{\mathrm{target}}^{\mathrm{current}}-L_{\mathrm{random}}^{\mathrm{current}}\bigr)
    -
    \bigl(L_{\mathrm{target}}^{\mathrm{superseded}}-L_{\mathrm{random}}^{\mathrm{superseded}}\bigr),
\]
subtracting the nonspecific effect of removing an equally long non-value span
within each validity condition. Positive values indicate that the same source
exerts greater influence on the current-value decision while its content
remains valid.

\begin{table}[t]
\centering
\caption{\textbf{Control-adjusted current-versus-superseded effect on the
binary current-value decision.} Difference-in-differences against a
token-length-matched random non-value span, on the same $214$ matched twin
pairs per checkpoint. Intervals are dialogue-level bootstraps
($2{,}000$ resamples, seed 954100); the twin pair is the dialogue.}
\label{tab:matched-validity-did}
\small
\begin{tabular}{lrr}
\toprule
Model & $D_{\mathrm{validity}}$ & 95\% CI \\
\midrule
Qwen-7B-IT             & $+.8598$ & $[+.8084, +.9065]$ \\
Llama-8B-IT           & $+.5607$ & $[+.4907, +.6308]$ \\
Mistral-7B-IT        & $+.6121$ & $[+.5421, +.6822]$ \\
OLMo-7B-IT         & $+.0514$ & $[+.0187, +.0888]$ \\
Mistral-24B-IT & $+.6916$ & $[+.6262, +.7523]$ \\
OLMo-32B-IT        & $+.7009$ & $[+.6355, +.7617]$ \\
Qwen-72B-IT            & $+.5187$ & $[+.4533, +.5841]$ \\
Llama-70B-IT          & $+.6589$ & $[+.5935, +.7196]$ \\
\midrule
Descriptive macro mean          & $+.5818$ & \\
\bottomrule
\end{tabular}
\end{table}

The adjusted effect is positive on all eight checkpoints and every interval
excludes zero (Table~\ref{tab:matched-validity-did}). Conversational validity
therefore modulates the same source's influence on the model's current-value
decision, and the difference is not explained by the generic effect of
removing an equally long span.
\subsection{Temporal Evolution of Original-Source Influence}
\label{app:temporal-source-influence}

At each available lag $\tau$ relative to the update, define the margin
\[
    z(H)=\log P(v_{\mathrm{old}}\mid H)
         -\log P(v_{\mathrm{new}}\mid H).
\]
We orient the two masking effects so that positive values indicate support
for the value at the masked source:
\begin{align}
 C_{\mathrm{old}} & = z(H)-z(H_{\mathrm{mask\ old}}),\\
 C_{\mathrm{new}} & = z(H_{\mathrm{mask\ new}})-z(H),\\
 D & = C_{\mathrm{new}}-C_{\mathrm{old}}.
\end{align}
Positive $C_{\mathrm{old}}$ means that the original old-value source supports
$v_{\mathrm{old}}$, positive $C_{\mathrm{new}}$ means that the original
new-value source supports $v_{\mathrm{new}}$, and positive $D$ indicates
stronger dependence on the new source.
The primary \emph{single-anchor} intervention blocks only the original
old-value mention and the new-value mention introduced at the update turn.
Later user or system restatements remain visible. These quantities therefore
measure dependence on the \emph{original source mentions}, not total use of
either value.
The model-independent paired sample sizes for $D$ are 727, 532, 498, 428,
and 334 at $\tau=0,+1,+2,+3,+4$, respectively. Each lag pairs the old- and
new-source interventions on eligible events. Because the eligible population
changes across lags, the aggregate trajectory is not a fixed-cohort estimate
of within-event decay.
\begin{table}[t]
\centering
\caption{\textbf{Relative dependence on original new and old source
mentions.} $D=C_{\mathrm{new}}-C_{\mathrm{old}}$ with 95\% bootstrap
intervals. Positive values indicate stronger dependence on the newly
introduced source.}
\label{tab:temporal-source-difference}

\scriptsize\setlength{\tabcolsep}{2pt}
\begin{tabular}{lccccc}
\toprule
Model & $\tau=0$ & $+1$ & $+2$ & $+3$ & $+4$ \\
\midrule
Qwen-7B-IT
& +16.62 [+14.5,+17.7]
& +1.50 [+.2,+2.7]
& +.85 [-.3,+2.0]
& +.00 [-1.2,+1.3]
& -.32 [-1.9,+1.1] \\
Llama-8B-IT
& +8.58 [+7.8,+9.4]
& -.23 [-.8,+.4]
& -.85 [-1.5,-.2]
& -1.13 [-1.8,-.4]
& -1.72 [-2.6,-.9] \\
Mistral-7B-IT
& +14.33 [+13.0,+15.6]
& +.60 [-.3,+1.5]
& -.76 [-1.6,+.1]
& -1.22 [-2.1,-.3]
& -1.81 [-2.8,-.8] \\
OLMo-7B-IT
& +8.42 [+7.4,+9.4]
& -1.68 [-2.4,-1.0]
& -2.02 [-2.8,-1.2]
& -2.86 [-3.7,-2.0]
& -2.99 [-4.0,-2.0] \\
Mistral-24B-IT
& +9.07 [+8.4,+9.7]
& +1.07 [+.6,+1.5]
& +.49 [+.0,+1.0]
& +.04 [-.5,+.6]
& -.14 [-.8,+.5] \\
OLMo-32B-IT
& +8.72 [+7.9,+9.6]
& -.14 [-.8,+.5]
& -.41 [-1.0,+.2]
& -.86 [-1.6,-.2]
& -1.06 [-1.8,-.2] \\
Qwen-72B-IT
& +19.21 [+17.5,+21.0]
& +3.14 [+1.9,+4.5]
& +1.68 [+.3,+3.1]
& +.93 [-.4,+2.3]
& +.46 [-1.2,+2.1] \\
Llama-70B-IT
& +9.28 [+8.4,+10.1]
& +1.41 [+.7,+2.1]
& +.36 [-.4,+1.1]
& -.12 [-.9,+.6]
& -.55 [-1.5,+.4] \\
\bottomrule
\end{tabular}
\end{table}
At the update turn, $D$ is positive with an interval excluding zero in all eight
checkpoints: dependence on the original newly introduced source immediately
exceeds dependence on the original old source. The estimated advantage is substantially smaller at later lags. By $\tau=+4$, the difference is no longer distinguishable from zero
for Qwen-7B-IT, Mistral-24B-IT, Qwen-72B-IT, and Llama-70B-IT.
For Llama-8B-IT, Mistral-7B-IT, OLMo-7B-IT, and OLMo-32B-IT, dependence on the original
old source is significantly stronger at that lag.
Importantly, this does not imply that the model ``returns'' to the old value
or stops using the new one. The intervention targets only the original
mentions, while later restatements remain available.
Even after the immediate new-source advantage contracts,
$C_{\mathrm{old}}$ remains substantial at $\tau=+4$, ranging from $+4.73$
to $+11.59$. Thus, the original old source continues to exert measurable
causal influence several turns after the update.
Matched random and unrelated controls remain small relative to these target
effects. At $\tau=+4$, random-old effects range from $-.088$ to $+.046$,
random-new from $-.071$ to $+.071$, and unrelated-old from $-.049$ to
$+.136$. Same-line controls are somewhat larger but remain far below the
magnitude of the historical-source effects.

\section{Natural Failure Diagnostics}
\label{app:natural-failures}
This appendix reports the complete diagnostic and intervention results of
Section~\ref{sec:natural}.
\subsection{Operational Failure Profiles}
\label{app:operational-profiles}
For each candidate slot at a natural SEARCH decision, we combine a calibrated
structural signal with a behavioral conditional-value margin.
The structural signal $S$ is obtained from the linear slot-presence readout
at the fixed decision layer (Appendix~\ref{app:readouts}). Readout scores are Platt-calibrated on
simulator development data. We define
\[
S=
\begin{cases}
\textsc{active}, & p \ge .70,\\
\textsc{inactive}, & p \le .30,\\
\textsc{ambiguous}, & \text{otherwise}.
\end{cases}
\]
The behavioral signal is a conditional value margin from the unmodified
model, not a readout. Let $v^*$ be the annotated current value and let
$P_s(v\mid H)$ denote the model's sequence probability for $v$ after the
canonical query prefix \texttt{domain: \{domain\}; \{slot\}:}. We compute
\[
 z=\log P_s(v^*\mid H)
    -\max_{v\in\mathcal{V}_{\mathrm{comp}}}\log P_s(v\mid H),
\]
where the candidates include superseded values, the emitted value, and the
highest-scoring database-vocabulary alternatives. Candidates equivalent to
the current value under the frozen normalization rules are excluded.

The threshold $\tau_z$ is selected on development data as the smallest value in
$\{0,.5,1,\ldots,8\}$ for which the sign of $z$ agrees with whether the emitted
value denotes the current value on at least $90\%$ of qualifying examples,
with at least 30 examples remaining. The realized threshold is $\tau_z=0$
for every checkpoint except OLMo-7B-IT ($\tau_z=2.5$).

The $.70/.30$ thresholds above define diagnostic profiles only. They are
not the controller's add/delete operating points. For natural omissions,
we assign profiles in the following priority order:
\[
\textsc{profile} =
\begin{cases}
\textsc{Structure}, &
S=\textsc{inactive},\\
\textsc{Ambiguous}, &
S=\textsc{ambiguous}
\ \text{or}\ z\ \text{undefined}
\ \text{or}\ |z|<\tau_z,\\
\textsc{Binding}, &
z<-\tau_z,\\
\textsc{Deployment}, &
\text{otherwise}.
\end{cases}
\]
Thus, \textsc{Structure} denotes weak structural support for the omitted slot,
\textsc{Binding} denotes unmodified-model scoring that favors a competing value,
\textsc{Deployment} denotes cases where the slot is structurally supported and
the current value is not decisively disfavored under the stated threshold rule despite omission, and
\textsc{Ambiguous} collects unresolved cases. The rule assigns structural
inactivity first, even if the value signal is undefined. At an exact margin
boundary, the strict inequalities above apply; in particular, $z=0$ with
$\tau_z=0$ falls in the residual \textsc{Deployment} category and should not
be interpreted as a strict preference for the current value.
\subsection{Profile Frequencies Across Checkpoints}
\label{app:profile-frequencies}
Profiles are assigned on the Base LM closed-loop trajectories of a separate fixed
500-goal simulator test set, disjoint from the controller's development and test goals,
with one label per offending slot. Calibration is performed on simulator DEV
and then applied unchanged. Seven checkpoints have complete profile results; Mistral-24B-IT does not.
\begin{table}[t]
\centering
\caption{\textbf{Natural omission profiles.}
Counts are per offending slot; fractions are relative to the omission count
for that checkpoint. Raw totals should not be compared across models because
the number of SEARCH decisions differs.}
\label{tab:omission-profiles}
\scriptsize
\setlength{\tabcolsep}{3pt}
\begin{tabular}{lrrrrrr}
\toprule
Model & SEARCH & Omissions &
Structure & Binding & Deployment & Ambiguous \\
\midrule
Qwen-7B-IT
& 2379 & 687
& 230 (.335) & 38 (.055) & 197 (.287) & 222 (.323) \\
Llama-8B-IT
& 2869 & 365
& 152 (.416) & 28 (.077) & 72 (.197) & 113 (.310) \\
Mistral-7B-IT
& 2631 & 173
& 61 (.353) & 11 (.064) & 19 (.110) & 82 (.474) \\
OLMo-7B-IT
& 2760 & 314
& 58 (.185) & 8 (.025) & 173 (.551) & 75 (.239) \\
OLMo-32B-IT
& 2165 & 133
& 55 (.414) & 5 (.038) & 21 (.158) & 52 (.391) \\
Qwen-72B-IT
& 2527 & 424
& 149 (.351) & 36 (.085) & 131 (.309) & 108 (.255) \\
Llama-70B-IT
& 2345 & 780
& 298 (.382) & 72 (.092) & 89 (.114) & 321 (.412) \\
\bottomrule
\end{tabular}
\end{table}
The profile mixture varies substantially across checkpoints. For example,
\textsc{Deployment} accounts for $.551$ of OLMo-7B-IT omissions but only $.110$
of Mistral-7B-IT omissions, while \textsc{Ambiguous} ranges from $.239$ to
$.474$. We therefore treat the profiles as a diagnostic stratification rather
than an exhaustive universal decomposition of failure.
For reference, the same frozen population also contains
\emph{extra-slot-only}, \emph{other-wrong-value-only}, and \emph{stale-value-only}
error categories. \emph{stale-value-only} is rare in most checkpoints, with only
1--29 examples, and is treated descriptively when the frozen population
contains fewer than 20 cases.
\subsection{Forcing an Omitted Slot}
\label{app:slot-forcing}
We next ask whether the profiles predict response to a targeted behavioral
intervention. Slot forcing preserves the Base LM query text generated so
far and appends only
\[
\texttt{; \{slot\}:}
\]
for the omitted slot. The model then freely generates at most 12 additional
tokens; generation is truncated at the first semicolon or newline and the
resulting query is reparsed through the standard executable-query pipeline.
These are fixed-prefix query interventions, not closed-loop controller runs.
For omission cases, repair succeeds only when the slot appears in the
executable query and the generated value denotes the gold current value.
Collateral damage measures whether other previously correct gold slots are
broken; new unsupported constraints are counted separately.
\begin{table}[t]
\centering
\caption{\textbf{Slot forcing on Qwen-7B-IT and Llama-8B-IT.}
Repair requires both insertion of the omitted slot and generation of the
current value. $\Delta$Exact and $\Delta$DB are relative to the Base LM query.
Collateral damage and new-extra counts are exactly zero in every displayed slot-forcing cell.}
\label{tab:slot-forcing-primary}
\scriptsize
\setlength{\tabcolsep}{3pt}
\begin{tabular}{llrrccc}
\toprule
Model & Profile & $n$ & Goals & Repair [95\% CI] &
$\Delta$Exact & $\Delta$DB \\
\midrule
Qwen-7B-IT
& Deployment & 197 & 81 & .934 [.884,.970] & +.579 & +.421 \\
& Structure  & 230 & 64 & .478 [.271,.672] & +.187 & +.070 \\
& Binding    & 38  & 20 & .132 [.033,.237] & +.053 & -.079 \\
& Ambiguous  & 222 & 99 & .554 [.420,.667] & +.302 & +.135 \\
\midrule
Llama-8B-IT
& Deployment & 72  & 46 & .486 [.348,.618] & +.333 & +.125 \\
& Structure  & 152 & 49 & .125 [.034,.214] & +.020 & -.092 \\
& Binding    & 28  & 20 & .036 [.000,.130] & +.036 & +.036 \\
& Ambiguous  & 113 & 53 & .133 [.027,.254] & +.062 & -.018 \\
\bottomrule
\end{tabular}
\end{table}
The strongest distinction is between \textsc{Deployment} and
\textsc{Structure}. Table~\ref{tab:slot-forcing-contrast} reports this contrast
for every checkpoint with a completed slot-forcing evaluation.
\begin{table}[t]
\centering
\caption{\textbf{Slot-forcing Deployment--Structure contrast.}
The 95\% interval for the repair-rate difference excludes zero in all six evaluated checkpoints, favoring
\textsc{Deployment} over \textsc{Structure}.}
\label{tab:slot-forcing-contrast}
\small
\begin{tabular}{lcc}
\toprule
Model & Deployment$-$Structure [95\% CI] & $n_D,n_S$ \\
\midrule
Qwen-7B-IT     & +.456 [+.275,+.681] & 197, 230 \\
Llama-8B-IT    & +.361 [+.224,+.508] & 72, 152 \\
Mistral-7B-IT  & +.602 [+.330,+.848] & 19, 61 \\
OLMo-7B-IT     & +.445 [+.337,+.547] & 173, 58 \\
OLMo-32B-IT    & +.807 [+.560,+.952] & 21, 55 \\
Qwen-72B-IT    & +.669 [+.478,+.811] & 131, 149 \\
\bottomrule
\end{tabular}
\end{table}
The contrast excludes zero in all six checkpoints. However, this should be
interpreted as \emph{predictive stratification}, not independent causal
identification of a deployment mechanism. The behavioral margin used to
construct the profile and slot forcing both explicitly condition on
the slot key, although they otherwise differ: the profile uses a canonical
two-field query prefix with teacher-forced candidates, whereas slot forcing retains
the Base LM draft and generates freely.
\subsection{Targeted Historical-Source Masking on Natural Failures}
\label{app:natural-stale-masking}
The profile analyses above are predictive. To obtain a more direct causal test
on naturally occurring errors, we isolate decisions in which the unmodified model
emits a superseded value after an explicit conversational update and mask the
historical source of that stale value.
This is an \emph{oracle-targeted, frozen-prefix intervention}: oracle
annotations are used only to identify the offending source span and to score
the resulting query. No oracle value or state is inserted into the prompt,
and the intervention does not continue the trajectory in closed loop.
We compare masking the superseded source against masking the currently valid
source, an unrelated value source, and a matched random non-value span.
Decisions come from a separate closed-loop run of the unmodified Base LM over
6,000 simulator goals for Qwen-7B-IT and Llama-8B-IT, whose first 500 are the
profile-analysis goals of Appendix~\ref{app:operational-profiles}. An update
event is eligible, before any output is inspected, when the user has
mentioned both the superseded and the current value of a slot inside the
visible context (753 and 884 events). The primary population contains every
eligible event in which the Base LM emits the superseded value and not the
current one (29 and 121); unlike the stale-value-only profile, other slots
may also be wrong. The correct-decision population contains eligible events
in which it emits only the current value (692 and 702).
\begin{table}[t]
\centering
\caption{\textbf{Natural stale-value intervention on the prespecified
Qwen-7B-IT and Llama-8B-IT populations.}
Repair means that the regenerated query emits the current value for the
updated slot. Qwen-7B-IT: 17/29 on all stale
errors; the main text's 17/28 uses the common paired population, where one
unrelated-control span is unavailable.}
\label{tab:natural-stale-primary}
\small
\setlength{\tabcolsep}{4pt}
\begin{tabular}{lrrrr}
\toprule
Model & Stale source & Current source & Unrelated & Random \\
\midrule
Qwen-7B-IT
& 17/29 (.586)
& 0/29 (.000)
& 3/28 (.107)
& 3/29 (.103) \\
Llama-8B-IT
& 47/121 (.388)
& 1/121 (.008)
& 9/121 (.074)
& 8/121 (.066) \\
\bottomrule
\end{tabular}
\end{table}
For Qwen-7B-IT, one unrelated-control span is unavailable, leaving a common
paired population of 28 examples. On that shared population, stale-source
masking repairs 17/28 ($.607$), compared with 3/28 ($.107$) for both random
and unrelated controls and 0/28 for masking the current source.
The same intervention improves DB equivalence in 13/29 Qwen cases and
34/121 Llama cases, with zero DB-equivalence losses in either model.
We also evaluate the interventions on decisions that are initially correct.
For Qwen-7B-IT, masking the currently valid source breaks 624/692 ($.902$) of
eligible correct decisions, whereas stale-source masking breaks only
2/692 ($.003$). On the 687-example common population the corresponding
figures are 620/687 ($.902$) and 2/687 ($.003$). For Llama-8B-IT,
current-source masking breaks 592/702 ($.843$), compared with 9/702 ($.013$)
under stale-source masking.
Thus, the same historical-source manipulation is strongly asymmetric:
suppressing an offending superseded source can correct some natural stale
decisions, whereas suppressing the currently valid source is highly
destructive. This supports a causal role for historical evidence in this
subset of natural failures, but does not imply that stale-history interference
accounts for natural failure in general. On four additional checkpoints, each run over 3,000 goals
(Mistral-7B-IT, OLMo-7B-IT,
Mistral-24B-IT, OLMo-32B-IT), the stale
populations are too small for inference ($n=9,8,8,5$), but stale-source
masking again repairs more cases than random or unrelated masking in each
($.222$--$.875$ against at most $.200$). Mistral-24B-IT has this replay but
no profile analysis in Appendix~\ref{app:operational-profiles}.

\section{State--Action Control: Ablations and Full Results}
\label{app:controller}

This appendix reports the ablations, operating points, and full closed-loop
results of the State--Action Controller (Section~\ref{sec:control}).

\subsection{Evidence Protection Ablation}
\label{app:d0-d1}

Evidence protection vetoes a proposed deletion when the latest explicit user
evidence supports a non-\texttt{dontcare} value for that slot. We isolate this
component at the conservative operating point $(.90,.30)$: the two arms use
the same structural thresholds, value-resolution rules, legality checks, and
compiler, differing only in the deletion veto.

\begin{table}[t]
\centering\scriptsize
\caption{\textbf{Evidence protection at the 90/30 operating point.}
Each arm evaluates 500 development goals. Destructive corrections, deletions,
and vetoes are counts; preservation and repair are within-turn conditional
rates defined in Appendix~\ref{app:controller-diagnostics}. Exact and DB-eq
are goal-weighted.}
\label{tab:evidence-protection}
\setlength{\tabcolsep}{3pt}
\begin{tabular}{llrrrrrrrrr}
\toprule
Model & Protection & Turns & Success & Exact & DB-eq & Preserve & Repair & Destructive & Deletions & Vetoes\\
\midrule
Qwen-7B-IT & Off & 2325 & .502 & .694 & .701 & .970 & .334 & 35 & 764 & 0\\
 & On & 2306 & .502 & .703 & .711 & 1.000 & .344 & 0 & 638 & 108\\
Llama-8B-IT & Off & 2875 & .206 & .602 & .619 & .928 & .314 & 83 & 1448 & 0\\
 & On & 2901 & .202 & .612 & .625 & 1.000 & .294 & 0 & 1216 & 287\\
\bottomrule
\end{tabular}
\end{table}

Protection eliminates the observed destructive corrections in both complete
audited development comparisons: 35 to zero for Qwen-7B-IT and
83 to zero for Llama-8B-IT. Exact-query accuracy and database
equivalence improve, while task success changes from $.502$ to $.502$ and
from $.206$ to $.202$, respectively. The paired success-difference intervals
are $[-.012,+.010]$ and $[-.016,+.008]$. Thus, the clearest benefit is
preserving correct queries rather than a resolved increase in task success.
The veto cancels 108 of 746 and 287 of 1,503 proposed deletions. All three
reported operating points retain this safeguard; these ablation results
should not be interpreted as a test-set ablation or as zero damage at every
operating point.

\subsection{Selector-Attribution Ablation}
\label{app:selector-attribution}

This ablation isolates the contribution of the learned structural readout. It
is an offline, fixed-prefix analysis over the logged Base LM SEARCH turns of
the $500$-goal development trajectories for Qwen-7B-IT
($2{,}405$ turns) and Llama-8B-IT ($2{,}949$ turns). The simulator
is not rerun and no new generation is performed: each arm reuses the same
dialogue history, the same Base LM draft, and the same cached decision-site
representation, so every arm is scored on exactly the same turns. Only the
source of the structural selection signal differs; the evidence resolver,
evidence protection, legality checks, compilation, and fallback are the
frozen controller code in every arm.

We compare the Base LM draft; full SAC at $.70/.50$; a resolver-only arm that
fixes $p_s=.5$ so no structural proposal can pass either threshold while the
deterministic resolution paths remain active; an observable selector that
replaces the readout with a ternary status derived only from the controller's
own visible-text evidence; a shuffled-readout arm that keeps the real
calibrated posterior vectors but permutes them jointly across turns of
different goals within the same domain; and a rate-matched random arm that
selects proposals uniformly at random while matching full SAC's proposal
counts per domain, edit type, and slot. The shuffled control changes the
intervention rate as well as its targeting, which is why the rate-matched
control is reported alongside it. The two random arms use $1{,}000$
permutations each (seeds $954100+k$); we report their means.

\begin{table}[t]
\centering\scriptsize
\caption{\textbf{Selector attribution: goal-weighted exact-query accuracy
under alternative selection signals.} Offline fixed-prefix replay on the
logged development SEARCH turns; all arms share the evidence resolver,
safeguards, compiler, and fallback. SAC$-$arm is the paired difference with a
goal-clustered bootstrap interval ($2{,}000$ resamples, seed 954100); for the
shuffled and rate-matched arms it uses the single seed-954100 draw, so it
differs slightly from the difference of the $1{,}000$-draw mean accuracies.}
\label{tab:selector-attribution}
\setlength{\tabcolsep}{3pt}
\begin{tabular}{lcccc}
\toprule
 & \multicolumn{2}{c}{Qwen-7B-IT} & \multicolumn{2}{c}{Llama-8B-IT} \\
\cmidrule(lr){2-3}\cmidrule(lr){4-5}
Selector & Exact & SAC$-$arm [95\% CI] & Exact & SAC$-$arm [95\% CI] \\
\midrule
Base LM draft             & .518 & -- & .430 & -- \\
Full SAC $.70/.50$        & \textbf{.708} & -- & \textbf{.640} & -- \\
Resolver only ($p_s=.5$)  & .519 & $+.190$ $[+.170,+.209]$ & .435 & $+.205$ $[+.179,+.230]$ \\
Observable text selector  & .501 & $+.208$ $[+.183,+.235]$ & .406 & $+.234$ $[+.207,+.262]$ \\
Shuffled readout          & .586 & $+.127$ $[+.106,+.148]$ & .509 & $+.130$ $[+.108,+.153]$ \\
Rate-matched random       & .532 & $+.174$ $[+.155,+.195]$ & .485 & $+.157$ $[+.134,+.180]$ \\
\bottomrule
\end{tabular}
\end{table}

Full SAC exceeds every alternative selector on both checkpoints, and all eight
intervals exclude zero. Disabling structural selection while retaining the
deterministic resolution paths recovers little over the Base LM draft
($.519$ and $.435$ against $.518$ and $.430$), and a selector built from
the controller's own visible-text evidence falls below the Base LM draft.

The edit-level diagnostics locate the difference. Full SAC's accepted-edit
precision is $.848$ and $.834$; its addition proposals reach precision and
recall $.626/.395$ and $.631/.402$, and its deletion proposals $.695/.675$
and $.687/.680$. The observable selector attains comparable addition
precision but far lower deletion recall, because an unsupported constraint is
by definition one the user never stated, so no matching text span exists and
absence of a span cannot be read as evidence of inactivity. The controller's
benefit therefore depends on dialogue-specific targeting by the learned
structural readout: deterministic value rules, observable evidence, shuffled
scores, and rate-matched random targeting do not reproduce it.

\subsection{Edit Composition and Legality-Only Rescoring}
\label{app:legality-filter}
Table~\ref{tab:legality-filter} decomposes the $21{,}591$ edits the primary
controller executes on TEST1000 and compares it with a legality-only
rescoring. The rescoring takes every logged Base LM SEARCH query, removes each
slot whose value is database-illegal or off-schema, adds nothing, and scores
the result against the same reference query; no trajectory is rerun. Of the
controller's edits, $12{,}641$ ($59\%$) are deletions, $98.5\%$ of which
target slots absent from the reference state; $4{,}556$ deletions ($21\%$ of
all edits) remove database-illegal values and the remaining $64\%$ of
deletions remove schema-legal ones. Additions account for $8{,}595$ edits and
value replacements for $355$. Legality-only rescoring raises five-checkpoint
exact-query accuracy from $.318$ to $.445$ and database equivalence from
$.379$ to $.552$, against $.621$ and $.640$ for the controller.

\begin{table}[t]
\centering
\caption{\textbf{Controller edit composition and legality-only rescoring on
TEST1000.} Edit counts are pooled over each checkpoint's SAC-P SEARCH turns.
Exact and DB-eq are goal-weighted over the logged Base LM SEARCH turns before
and after removing illegal slots.}
\label{tab:legality-filter}

\scriptsize\setlength{\tabcolsep}{2pt}
\begin{tabular}{lrrrrrrrr}
\toprule
& \multicolumn{5}{c}{SAC-P edits} & \multicolumn{3}{c}{Legality-only rescoring} \\
\cmidrule(lr){2-6}\cmidrule(lr){7-9}
Model & Edits & Deletions & Illegal del. & Additions & Replacements & Turns changed & Exact & DB-eq \\
\midrule
Qwen-7B-IT      & 2{,}284 & 1{,}365 & 371     & 897     & 22  & .166 & .487$\to$.545 & .567$\to$.656 \\
Llama-8B-IT    & 3{,}065 & 1{,}999 & 947     & 938     & 128 & .400 & .389$\to$.501 & .442$\to$.617 \\
Mistral-7B-IT & 4{,}507 & 3{,}891 & 2{,}007 & 504     & 112 & .752 & .165$\to$.449 & .205$\to$.581 \\
OLMo-7B-IT  & 9{,}107 & 3{,}091 & 756     & 5{,}960 & 56  & .462 & .068$\to$.118 & .096$\to$.167 \\
OLMo-32B-IT & 2{,}628 & 2{,}295 & 475     & 296     & 37  & .200 & .482$\to$.611 & .586$\to$.741 \\
\midrule
Total / mean             & 21{,}591 & 12{,}641 & 4{,}556 & 8{,}595 & 355 & .396 & .318$\to$.445 & .379$\to$.552 \\
\bottomrule
\end{tabular}
\end{table}

\subsection{Controller Rescue by Natural-Omission Profile}
\label{app:profile-rescue}
We replay the frozen $.70/.50$ controller offline on exactly the omissions
profiled in Section~\ref{sec:natural} (Table~\ref{tab:omission-profiles}),
reusing each logged Base LM draft, visible history, and cached decision-site
activation; no language-model call is made. An omission is rescued when the
controller's query contains the omitted slot with its current value. Profiles
and controller share the calibrated slot readout, so no \textsc{Structure}
omission can be proposed, and only $6$ of $973$ \textsc{Ambiguous} omissions
(all on OLMo-7B-IT) receive one; the informative contrast is \textsc{Binding} versus
\textsc{Deployment}, where every slot is proposed. Of the $198$ pooled
\textsc{Binding} omissions, $26$ are rescued, $112$ receive no usable value
because the visible user turns contain none, and $60$ receive a different
value. Mistral-24B-IT has no profile results.

\begin{table}[t]
\centering
\caption{\textbf{Controller rescue by natural-omission profile.} Offline
fixed-prefix replay on the profiled omissions; rescue is the fraction of
omissions whose controller query contains the slot with its current value.
Intervals are 95\% bootstrap intervals over omissions.}
\label{tab:profile-rescue}
\scriptsize
\begin{tabular}{lrcrc}
\toprule
& \multicolumn{2}{c}{\textsc{Binding}} & \multicolumn{2}{c}{\textsc{Deployment}} \\
\cmidrule(lr){2-3}\cmidrule(lr){4-5}
Model & $n$ & Rescue & $n$ & Rescue \\
\midrule
Qwen-7B-IT & 38 & .158 [.053,.289] & 197 & .863 [.817,.909] \\
Llama-8B-IT & 28 & .143 [.036,.286] & 72 & .583 [.472,.694] \\
Mistral-7B-IT & 11 & .273 [.000,.545] & 19 & .947 [.842,1.00] \\
OLMo-7B-IT & 8 & .125 [.000,.375] & 173 & .988 [.971,1.00] \\
OLMo-32B-IT & 5 & .600 [.200,1.00] & 21 & .714 [.524,.905] \\
Qwen-72B-IT & 36 & .083 [.000,.167] & 131 & .702 [.618,.779] \\
Llama-70B-IT & 72 & .083 [.028,.153] & 89 & .865 [.797,.933] \\
\midrule
Pooled & 198 & .131 & 702 & .833 \\
\bottomrule
\end{tabular}
\end{table}

\subsection{Exact Controller Procedure}
\label{app:d1-exact}

Let $q_{\mathrm{base}}$ be the Base LM query, $K_t$ its executable slot
keys, and $E_t$ the domain-valid constrained slots covered by the readout.
The calibrated posterior $p_s$ is read once from the fixed decision-layer
representation. Structural proposals are
\begin{align}
\mathcal A_t &=\{s\in E_t\setminus K_t:p_s>\tau_{\mathrm{add}}\},\\
\mathcal D_t &=\{s\in K_t:p_s<\tau_{\mathrm{del}}\}.
\end{align}
The controller inherits the parsed Base LM domain. An invalid or unsupported
domain causes abstention. Slots between the thresholds retain their original
presence decision, although an evidence-supported value update can still
apply to a retained slot.

\begin{enumerate}
\item \textbf{Protect.} For each proposed deletion, inspect the visible user
evidence for that domain and slot. If its latest item is not
\texttt{dontcare}, cancel the deletion.
\item \textbf{Resolve.} For each added slot, use the latest explicit
user-supported value; absent evidence or a latest \texttt{dontcare} value
prevents the addition. For a retained Base LM slot, replace its value only
when a different explicit value occurs strictly later than the evidence
supporting the emitted value and its utterance contains a fixed update cue.
Otherwise retain the emitted value.
\item \textbf{Validate.} Check each proposed value against the domain's
ontology and database-legality rules. An illegal replacement falls back to
the Base LM value; an illegal addition is omitted. Compile in canonical
slot order and reparse the entire query. Execute the edit only if reparsing
recovers the intended domain and fields; otherwise retain the Base LM query.
\end{enumerate}

The procedure acts on the generated query and available dialogue text. It
maintains no persistent complete slot--value state and introduces no
additional language-model generation. Evidence resolution, protection, and
validation are identical across operating points.

\paragraph{Evidence extraction.}
\emph{Explicit user evidence} for a domain--slot pair is the ordered list of
values stated in the user's own turns within the rendered dialogue window
(at most 64 lines; Appendix~\ref{app:closed-loop-protocol}); lines that are
not user utterances are skipped. Candidate values are matched against a value
vocabulary derived from the MultiWOZ database records for that domain and
slot, extended by frozen surface-form alias maps, using whole-word matching
in which the longest match wins and nested shorter matches are discarded.
Four slots use dedicated frozen rules: a digit preceding ``star'' for
\texttt{stars}, a slot keyword together with a preceding negation cue for
\texttt{parking} and \texttt{internet}, and the longest canonical database
name for \texttt{name}. An explicit ``do not care'' together with the slot
word yields a \textsc{dontcare}. A frozen update-cue pattern marks sentences
that signal revision. All matched values are canonicalized before use.

The controller therefore reads only the visible user utterance text, the
domain parsed from the Base LM's own draft query, the database-derived value
vocabulary, and these frozen rules; the database is additionally used to
validate and execute the compiled query. It does not consult the simulator
goal, the reference belief state, the reference query, the rule-based state
tracker, the system-side NLU, system turns, future turns, or any hidden
simulator annotation. Because the primary closed-loop simulator uses template NLG, surface
variation is limited; transfer of the frozen matcher to human-authored
dialogue and naturalized user language is evaluated in
Appendices~\ref{app:human-transfer} and~\ref{app:naturalized}.

\subsection{Development Operating Points}
\label{app:controller-selection}

We report conservative $(.90,.30)$, intermediate $(.70,.50)$, and binary
$(.50,.50)$ settings, ordered as
$(\tau_{\mathrm{add}},\tau_{\mathrm{del}})$. Lowering the addition threshold
and raising the deletion threshold reduces structural abstention. At the
binary setting there is no positive-width abstention region, although an
exact $.50$ tie retains the original decision and all evidence safeguards
remain active.

The complete development comparison uses 500 goals each for
Qwen-7B-IT and Llama-8B-IT. The original development runs
preceded test evaluation; subsequent reruns reproduce these measurements.
We use 70/50 as the primary operating point for query fidelity and retain
90/30 and 50/50 as sensitivity endpoints. Among these three settings, 70/50
has the highest two-checkpoint mean exact-query accuracy ($.7075$) and
database equivalence ($.7065$), compared with $.6575$/$.6680$ at 90/30
and $.6930$/$.6895$ at 50/50. This is a query-fidelity rationale, not a
claim that 70/50 maximizes development task success.

\begin{table}[t]
\centering\scriptsize
\caption{\textbf{Complete two-checkpoint development comparison.}
All rows use 500 goals. Success is goal-level; Exact and DB-eq average over
SEARCH turns within each goal and then over goals. Preservation and repair
condition on draft correctness within the same turn. Rescue and destructive
counts are divided by goals; coverage is the fraction of SEARCH turns whose
executed query text differs from the Base LM draft.}
\label{tab:controller-dev-operating-points}
\setlength{\tabcolsep}{3pt}
\begin{tabular}{llrrrrrrrrr}
\toprule
Model & Setting & Turns & Success & Exact & DB-eq & Preserve & Repair & Destr./goal & Rescue/goal & Coverage\\
\midrule
Qwen-7B-IT & Base LM & 2405 & .486 & .518 & .599 & 1.000 & .000 & .000 & .000 & .000\\
 & 90/30 & 2306 & .502 & .703 & .711 & 1.000 & .344 & .000 & .782 & .302\\
 & 70/50 & 2280 & .490 & .753 & .745 & .955 & .468 & .108 & 1.010 & .405\\
 & 50/50 & 2238 & .490 & .741 & .731 & .926 & .489 & .176 & 1.028 & .445\\
\midrule
Llama-8B-IT & Base LM & 2949 & .156 & .430 & .466 & 1.000 & .000 & .000 & .000 & .000\\
 & 90/30 & 2901 & .202 & .612 & .625 & 1.000 & .294 & .000 & 1.034 & .498\\
 & 70/50 & 2865 & .214 & .662 & .668 & .957 & .395 & .098 & 1.366 & .580\\
 & 50/50 & 2831 & .206 & .645 & .648 & .901 & .406 & .230 & 1.358 & .604\\
\bottomrule
\end{tabular}
\end{table}

Coverage and repair increase and preservation decreases across the three
settings. The intermediate setting achieves the highest exactness and
database equivalence in both checkpoints among the three reported settings.
Task success is less discriminating and favors different settings across
checkpoints.

\subsection{Five-Checkpoint Closed-Loop Results}
\label{app:d1-test-details}

The shared test panel contains five checkpoints with all eight arms completed
on 1,000 initial goals each: Base LM, Prompted DST at $k\in\{0,5,10\}$, T5-DST,
and the three controller operating points. Coverage is documented in
Appendix~\ref{app:coverage}. Each arm interacts independently after its first
different query. Terminal success is paired by initial goal, while query
metrics describe each arm's own resulting trajectories
(Table~\ref{tab:controller-test-full}).

\begin{table}[t]
\centering
\caption{\textbf{Closed-loop task success and exact-query accuracy.}
Cells report Success/Exact on 1,000 goals per checkpoint and arm. Macro
results give each checkpoint equal weight and are computed before rounding.
All controller settings use the same evidence safeguards. PDST-$k$ is Prompted DST with $k$ exemplars; Oracle is the
privileged Oracle Query reference of Table~\ref{tab:control-main}.}
\label{tab:controller-test-full}

\scriptsize\setlength{\tabcolsep}{2pt}
\begin{tabular}{lccccccccc}
\toprule
Model & Base & PDST-0 & PDST-5 & PDST-10 & T5-DST & SAC-C & SAC-P & SAC-B & Oracle\\
\midrule
Qwen-7B-IT & .415/.487 & .437/.449 & .446/.396 & .454/.379 & .456/.574 & .431/.675 & .449/.719 & .433/.698 & .503/.999\\
Llama-8B-IT & .151/.389 & .172/.395 & .180/.430 & .194/.444 & .207/.483 & .184/.563 & .210/.608 & .180/.580 & .223/.938\\
Mistral-7B-IT & .086/.165 & .175/.403 & .150/.378 & .160/.382 & .185/.546 & .174/.447 & .181/.540 & .174/.514 & .226/1.000\\
OLMo-7B-IT & .130/.068 & .292/.190 & .359/.258 & .363/.277 & .426/.580 & .312/.301 & .376/.491 & .394/.569 & .471/.977\\
OLMo-32B-IT & .577/.482 & .563/.330 & .606/.459 & .644/.484 & .584/.563 & .644/.683 & .640/.745 & .620/.723 & .690/1.000\\
\midrule
Mean & .272/.318 & .328/.353 & .348/.384 & .363/.393 & .372/.549 & .349/.534 & .371/.621 & .360/.617 & .423/.983\\
\bottomrule
\end{tabular}
\end{table}

The primary controller improves exact-query accuracy over both the Base LM
and Prompted DST in every checkpoint. Few-shot prompting narrows the success
gap to $+.008$ at $k{=}10$ while the exact-query gap remains $+.227$
(Table~\ref{tab:paired-contrasts-fewshot}). Against T5-DST,
terminal-success differences are heterogeneous: SAC-P is higher on
OLMo-32B-IT ($+.056$ $[+.029,+.085]$, Holm $p=.0007$) and lower on
OLMo-7B-IT ($-.050$ $[-.074,-.025]$, Holm $p=.0004$), while the intervals for
Qwen-7B-IT, Llama-8B-IT, and Mistral-7B-IT include zero. Thus, the fixed-panel mean should
not be interpreted as uniform checkpoint-level dominance.

\paragraph{Paired uncertainty.}
Success differences use 2,000 goal-paired bootstrap resamples with seed
954100; exact McNemar tests use discordant goal outcomes. For example,
Qwen-7B-IT improves by $+.034$ at 70/50
(95\% CI $[+.014,+.054]$, $p=.0012$), versus $+.016$ at 90/30
($[-.002,+.034]$, $p=.0805$). For OLMo-32B-IT, differences
are $+.067$ ($[+.044,+.091]$), $+.063$ ($[+.038,+.089]$), and
$+.043$ ($[+.015,+.071]$) for 90/30, 70/50, and 50/50, respectively.
Table~\ref{tab:paired-contrasts} reports the primary controller's paired
task-success contrast against both the Base LM and Prompted DST for every
checkpoint, together with the paired exact-query contrast against Prompted
DST.
\begin{table}[t]
\centering
\caption{\textbf{Paired contrasts for the primary controller.}
Success differences are goal-paired on the shared $1{,}000$ initial goals per
checkpoint; intervals use $2{,}000$ paired bootstrap resamples (seed 954100)
and $p$-values are raw exact two-sided McNemar tests on discordant goals.
$^{*}$ marks comparisons that remain significant at $.05$ after Holm
correction across the five checkpoints within that comparator family; the
two families are corrected separately. \emph{Exact query} is goal-weighted
turn-level exact-query accuracy: each goal contributes the mean over its own
SEARCH turns, and goals are averaged with equal weight. Because edited queries
change downstream turns, the two arms' SEARCH turns are not matched, so this
column is paired only by initial goal and no McNemar test is reported for it.
The macro row is a fixed-panel contrast over the five checkpoints.}
\label{tab:paired-contrasts}

\scriptsize\setlength{\tabcolsep}{2pt}
\begin{tabular}{lccccc}
\toprule
 & \multicolumn{2}{c}{TS, SAC-P $-$ Base LM}
 & \multicolumn{2}{c}{TS, SAC-P $-$ Prompted DST}
 & Exact, $-$ Prompted\\
\cmidrule(lr){2-3}\cmidrule(lr){4-5}\cmidrule(lr){6-6}
Model & $\Delta$ [95\% CI] & $p$ & $\Delta$ [95\% CI] & $p$ & $\Delta$ [95\% CI]\\
\midrule
Qwen-7B-IT & $+.034$ $[+.014,+.054]$ & $.0012^{*}$ & $+.012$ $[-.010,+.033]$ & $.3112$ & $+.270$ $[+.250,+.290]$ \\
Llama-8B-IT & $+.059$ $[+.038,+.080]$ & $<.0001^{*}$ & $+.038$ $[+.016,+.059]$ & $.0007^{*}$ & $+.213$ $[+.192,+.235]$ \\
Mistral-7B-IT & $+.095$ $[+.073,+.117]$ & $<.0001^{*}$ & $+.006$ $[-.015,+.027]$ & $.6427$ & $+.137$ $[+.113,+.162]$ \\
OLMo-7B-IT & $+.246$ $[+.216,+.277]$ & $<.0001^{*}$ & $+.084$ $[+.051,+.114]$ & $<.0001^{*}$ & $+.301$ $[+.277,+.323]$ \\
OLMo-32B-IT & $+.063$ $[+.038,+.089]$ & $<.0001^{*}$ & $+.077$ $[+.047,+.107]$ & $<.0001^{*}$ & $+.416$ $[+.396,+.436]$ \\
\midrule
Macro & $+.099$ $[+.089,+.111]$ & & $+.043$ $[+.032,+.055]$ & & $+.267$ $[+.258,+.277]$ \\
\bottomrule
\end{tabular}
\end{table}

\begin{table}[t]
\centering
\caption{\textbf{Paired contrasts against few-shot Prompted DST.}
Same protocol as Table~\ref{tab:paired-contrasts}: success is goal-paired on
the shared $1{,}000$ initial goals per checkpoint with $2{,}000$ paired
bootstrap resamples (seed 954100) and raw exact two-sided McNemar
$p$-values; $^{*}$ marks comparisons significant at $.05$ after Holm
correction across the five checkpoints within that comparator family
(OLMo-32B-IT at $k{=}5$ has Holm $p=.053$). Exact-query differences are
goal-weighted and paired only by initial goal.}
\label{tab:paired-contrasts-fewshot}

\scriptsize\setlength{\tabcolsep}{1.2pt}
\begin{tabular}{lcccccc}
\toprule
 & \multicolumn{3}{c}{SAC-P $-$ Prompted DST ($k{=}5$)}
 & \multicolumn{3}{c}{SAC-P $-$ Prompted DST ($k{=}10$)}\\
\cmidrule(lr){2-4}\cmidrule(lr){5-7}
Model & TS $\Delta$ [95\% CI] & $p$ & Exact $\Delta$ [95\% CI]
      & TS $\Delta$ [95\% CI] & $p$ & Exact $\Delta$ [95\% CI]\\
\midrule
Qwen-7B-IT      & $+.003$ $[-.019,+.025]$ & $.856$ & $+.323$ $[+.303,+.343]$ & $-.005$ $[-.028,+.017]$ & $.723$ & $+.340$ $[+.318,+.359]$ \\
Llama-8B-IT    & $+.030$ $[+.008,+.052]$ & $.0079^{*}$ & $+.179$ $[+.159,+.199]$ & $+.016$ $[-.006,+.036]$ & $.171$ & $+.164$ $[+.144,+.184]$ \\
Mistral-7B-IT & $+.031$ $[+.008,+.052]$ & $.0080^{*}$ & $+.162$ $[+.137,+.185]$ & $+.021$ $[-.001,+.042]$ & $.078$ & $+.158$ $[+.132,+.181]$ \\
OLMo-7B-IT  & $+.017$ $[-.011,+.046]$ & $.273$ & $+.233$ $[+.213,+.254]$ & $+.013$ $[-.014,+.040]$ & $.395$ & $+.214$ $[+.194,+.237]$ \\
OLMo-32B-IT & $+.034$ $[+.008,+.061]$ & $.018$ & $+.286$ $[+.266,+.306]$ & $-.004$ $[-.030,+.021]$ & $.821$ & $+.261$ $[+.240,+.281]$ \\
\midrule
Macro & $+.023$ $[+.012,+.034]$ & & $+.237$ $[+.226,+.247]$ & $+.008$ $[-.002,+.019]$ & & $+.227$ $[+.217,+.238]$ \\
\bottomrule
\end{tabular}
\end{table}

\subsection{Preservation, Repair, and Destructive Corrections}
\label{app:controller-diagnostics}

At each SEARCH turn on an arm's own trajectory, let $b_t$ indicate whether
the Base LM draft is exact and $e_t$ whether the executed query is exact.
We count
\begin{align}
 N_{\mathrm{preserved}}&=\sum_t\mathbf{1}[b_t=1,e_t=1],&
 N_{\mathrm{destructive}}&=\sum_t\mathbf{1}[b_t=1,e_t=0],\\
 N_{\mathrm{rescue}}&=\sum_t\mathbf{1}[b_t=0,e_t=1],&
 N_{\mathrm{wrong}}&=\sum_t\mathbf{1}[b_t=0,e_t=0].
\end{align}
Preservation and repair are
\[
 P_{\mathrm{preserve}}=
 \frac{N_{\mathrm{preserved}}}{N_{\mathrm{preserved}}+N_{\mathrm{destructive}}},
 \qquad
 P_{\mathrm{repair}}=
 \frac{N_{\mathrm{rescue}}}{N_{\mathrm{rescue}}+N_{\mathrm{wrong}}}.
\]
These compare draft and execution on the \emph{same turn}, not turn $t$
against a different arm's turn $t$. The latter comparison is invalid once
trajectories diverge. For Base LM, draft and execution coincide, so
preservation is one and rescue and destruction are zero by construction.
Coverage counts textual changes to the query and need not equal the rate of
database-result changes. Table~\ref{tab:control-main} reports their
five-checkpoint means.

Moving from 90/30 to 50/50 increases both repair and destruction in the macro
results; 70/50 combines higher preservation than the binary setting with
higher repair than the conservative setting. Because arms visit different histories, these
conditional rates characterize the complete deployed interfaces rather than
an intervention effect on one shared fixed-prefix population.

\subsection{Slot-Level Fidelity Metric}
\label{app:slot-comparison}

A true positive requires the executable query and reference to contain the
same slot with an equivalent value. An emitted constraint absent from the
reference, or carrying a different value, is a false positive; an absent or
incorrectly valued reference constraint is a false negative. Thus, a wrong
value can contribute both a false positive and a false negative. Slot
precision, recall, and F1 pool slot instances over turns within a checkpoint.
They are distinct from dialogue-level inform precision and recall.

\subsection{Inference Requirements, Model Footprint, and Supervision}
\label{app:d1-cost}

The three interfaces differ in what they require at inference time. The
controller operates on the existing decision-layer representation and the
Base LM's already-generated query, adding no generative call, no generated
tokens, and no extra prefill. Prompted DST performs an additional
state-generation call with the same backbone at every SEARCH decision,
with $5.02$, $5.04$, and $4.99$ extra calls and $3{,}135$, $4{,}720$, and
$6{,}113$ additional prefill tokens per goal at $k{=}0,5,10$.
T5-DST invokes a separately trained encoder--decoder at every SEARCH
decision.

T5-DST averages $4.98$ separate-model calls per goal ($4.68$--$5.52$ across
checkpoints); because these do not pass through the backbone, a
backbone-only counter would miss them entirely.

\begin{table}[t]
\centering\small
\caption{\textbf{Measured per-decision latency (ms) and added peak memory.}
One NVIDIA A40, batch size 1, \texttt{bfloat16} backbone (T5-DST in float32), $300$ identical logged SEARCH decisions
from the TEST1000 Base LM trajectories per checkpoint; wall time between
device synchronizations. Added memory is peak allocated memory beyond the
resident backbone ($14.8$/$15.6$\,GiB). The Base LM draft is the shared
reference that every arm starts from; the other rows are additional work.
The controller row includes the readout, evidence resolution, and
compilation. On an H200 all rows are $2$--$4\times$ faster with the same
ordering.}
\label{tab:controller-latency}
\begin{tabular}{lrrrrl}
\toprule
& \multicolumn{2}{c}{Qwen-7B-IT} & \multicolumn{2}{c}{Llama-8B-IT} & \\
\cmidrule(lr){2-3}\cmidrule(lr){4-5}
Arm & median & p95 & median & p95 & Added memory\\
\midrule
Base LM draft (reference) & 655 & 1{,}148 & 1{,}202 & 2{,}002 & 1.9 / 4.5 GB\\
Prompted DST (0-shot) & 813 & 1{,}224 & 1{,}222 & 2{,}171 & 0.8 / 1.9 GB\\
Prompted DST ($k{=}5$) & 922 & 1{,}487 & 1{,}156 & 2{,}219 & 0.9 / 2.0 GB\\
Prompted DST ($k{=}10$) & 983 & 1{,}596 & 1{,}249 & 2{,}228 & 1.0 / 2.1 GB\\
T5-DST & 95 & 173 & 101 & 168 & 40 MB\\
SAC-P & 5.1 & 19 & 6.5 & 19 & 8 MB\\
\bottomrule
\end{tabular}
\end{table}

The controller's readout has parameter count $8d+8+2d$ at model width $d$,
plus two calibration scalars. Each readout is fit on $22{,}083$ training and
$2{,}441$ development items with LBFGS (learning rate 1, at most 200
iterations, strong-Wolfe line search); calibration uses $1{,}990$ development
turns containing $9{,}373$ slot instances. T5-small is fine-tuned on the same
$22{,}083$ decision sites from $5{,}932$ training dialogues, with one trained
tracker shared across all backbones (Appendix~\ref{app:t5-dst}).

These interfaces differ in supervision, objective, and model reuse rather
than in efficiency alone: T5-DST is an explicit-state alternative that
carries a complete predicted state on the inference path, whereas the
controller learns only a slot-presence signal and reuses the Base LM's own
action. The controller's strongest advantages are executable-query fidelity
and preservation of correct drafts; explicit state reconstruction remains a
competitive alternative for maximizing terminal task success.

\subsection{Transfer to Human-Authored Dialogue}
\label{app:human-transfer}

The closed-loop experiments in Section~\ref{sec:control} use ConvLab's
agenda-based user simulator with template user NLG. We therefore conduct a
separate fixed-history replay to test whether the frozen State--Action
Controller transfers to human-authored dialogue text. This experiment is
restricted to the Base LM and the primary controller operating point
$(\tau_{\mathrm{add}},\tau_{\mathrm{del}})=(.70,.50)$ and is intended as a
Base-relative transfer test rather than a closed-loop task-success comparison.

\paragraph{Population and protocol.}
We replay 4,490 matched turns drawn from 911 MultiWOZ~2.2 test dialogues:
1,588 hotel turns, 1,668 restaurant turns, and 1,234 attraction turns.
Starting from the 7,363 user turns in the test split, we exclude 1,187 turns
with no executable-domain constraint and 1,686 turns whose reference state
contains constraints from two or more executable domains. The latter
restriction matches the controller interface, in which one SEARCH query names
one domain.

Unlike the closed-loop simulator evaluation, generated queries are not
executed to determine subsequent dialogue turns. Base LM and SAC therefore
receive byte-identical dialogue histories at every evaluated decision, making
the query-level comparison exactly paired. Terminal task success is not
defined in this replay setting.

\begin{table}[t]
\centering
\caption{\textbf{Transfer to human-authored MultiWOZ dialogue.}
Base LM and the frozen primary controller are evaluated on 4,490 matched
turns from 911 test dialogues. Intervals use 2,000 dialogue-clustered
bootstrap resamples (seed 954100). This is fixed-history query evaluation;
terminal task success is undefined.}
\label{tab:human-transfer}
\small
\begin{tabular}{lccc}
\toprule
Metric & Base LM & SAC-P & SAC-P $-$ Base \\
\midrule
Exact-query accuracy
    & .168 & .326 & $+.1575$ $[+.1453,+.1698]$ \\
Database equivalence
    & .267 & .429 & $+.1623$ $[+.1499,+.1750]$ \\
Initially correct Base query broken
    & -- & .0097 $[.0072,.0126]$ & -- \\
Fraction of turns edited
    & -- & .5944 $[.5818,.6068]$ & -- \\
\bottomrule
\end{tabular}
\end{table}

\paragraph{Base-relative transfer.}
Across the five-checkpoint panel, SAC improves exact-query accuracy over its
own Base LM by $+.1575$ and database equivalence by $+.1623$. Both
dialogue-clustered bootstrap intervals exclude zero. At the same time, the
controller changes an initially correct Base query into an incorrect one on
only .0097 of evaluated turns. Thus, the selective-correction effect observed
in closed-loop simulation persists when the dialogue history is
human-authored, although the Base-relative query-fidelity gain is smaller than
in the template-simulator setting.

\paragraph{Slot-level behavior.}
At the slot level, SAC increases macro slot precision from .539 to
.766 and recall from .538 to .641. Unsupported constraints fall from 1.163 to
.533 per turn, while missing constraints fall from 1.251 to .971. The
improvement is therefore substantially larger on precision than recall,
consistent with a selective controller that is particularly effective at
removing unsupported constraints but remains limited in recovering missing
ones.

\paragraph{Evidence availability.}
The value resolver deliberately uses explicit evidence from user turns only.
Across 12,151 reference constraint values, it identifies the correct latest
value for 8,740 (.719), while 2,573 (.212) have no user-side evidence.
Coverage is lowest for \texttt{name} (.344): of 1,576 missed names, 1,014
(.643) appear only in system turns, as when a user accepts a system
recommendation with ``book it''. Accordingly, SAC's exact-query gain is
$+.275$ on turns whose reference values are all user-stated, against
$+.038$ on mixed-evidence turns and $+.079$ when values are system-side
(reweighted to a common slot-count distribution). A development-only resolver
with name canonicalization and fuzzy matching raises coverage to .755 but
end-to-end exact-query accuracy by only about .008, so the frozen resolver is
retained.

\paragraph{Scope.}
The replay is deliberately restricted to turns with at most one constrained
executable domain because the deployed SEARCH interface emits one domain per
query. The excluded multi-domain population (1,686 of 7,363 test turns) is
outside the controller's one-domain-per-query interface.

\subsection{Closed-Loop Evaluation with Naturalized User Language}
\label{app:naturalized}

The closed-loop protocol of Appendix~\ref{app:closed-loop-protocol} renders
user turns with ConvLab's template NLG, so the simulated user restates each
constraint in a fixed surface form. To vary the surface form while keeping
task success measurable, we rerun the Base LM and the primary controller on
the same $1{,}000$ TEST goals with every user utterance rewritten by
\texttt{gpt-4o-mini-2024-07-18} at temperature 0 before the agent sees it.
The agenda policy, its goal and act sequence, the database, the system-side
NLU, the reference state tracker, the evaluator, the agent prompts, and the
frozen readout and controller are unchanged; the evaluator and reference
tracker still consume the original user act, so goal semantics are untouched.
This is naturalized simulator language, not human dialogue: the user's intent
remains the agenda policy.

A faithfulness gate requires every slot value conveyed by the user act to
survive canonicalization in the rewrite; a failing rewrite is retried once and
otherwise replaced by the template utterance, and the run records the
fallback rate. Rewrites are cached by the utterance, its act payload, the
prompt version, and the model snapshot, so identical template turns receive
identical rewrites in both arms. Across the ten arm--model runs,
51,112 user turns required 14,799 API calls and fell back to the
template on 84 turns (0.16\%), with no API errors.

\begin{table}[t]
\centering
\caption{\textbf{Closed-loop results with naturalized user language.}
Both arms run on the $1{,}000$ TEST goals per checkpoint with rewritten user
utterances. Contrasts are goal-paired with $2{,}000$ bootstrap resamples (seed
954100); the Nat.$-$templ.\ contrast compares SAC-P under naturalized and template
language on the same goals, and is not significant on any checkpoint after
Holm correction. Fallback rate is the larger of the two arms' template
fallback rates.}
\label{tab:naturalized}

\scriptsize\setlength{\tabcolsep}{1.5pt}
\begin{tabular}{lccccccc}
\toprule
 & Base LM & SAC-P & SAC-P & SAC-P$-$Base & SAC-P$-$Base & Nat.$-$templ. & Fallback \\
Model & TS/Exact & TS/Exact & Preserv. & TS $\Delta$ [95\% CI] & Exact $\Delta$ & TS $\Delta$ [95\% CI] & rate \\
\midrule
Qwen-7B-IT & .399/.446 & .440/.706 & .955 & $+.041$ $[+.020,+.063]$ & $+.260$ & $-.009$ $[-.033,+.015]$ & .002 \\
Llama-8B-IT & .167/.354 & .186/.553 & .928 & $+.019$ $[-.003,+.041]$ & $+.198$ & $-.024$ $[-.050,+.001]$ & .002 \\
Mistral-7B-IT & .081/.149 & .189/.558 & .986 & $+.108$ $[+.087,+.132]$ & $+.409$ & $+.008$ $[-.016,+.032]$ & .002 \\
OLMo-7B-IT & .128/.058 & .370/.518 & .972 & $+.242$ $[+.208,+.275]$ & $+.461$ & $-.006$ $[-.036,+.024]$ & .004 \\
OLMo-32B-IT & .530/.426 & .619/.739 & .977 & $+.089$ $[+.061,+.118]$ & $+.314$ & $-.021$ $[-.051,+.008]$ & .002 \\
\midrule
Mean & .261/.287 & .361/.615 & .964 & $+.100$ $[+.087,+.111]$ & $+.328$ & $-.010$ $[-.023,+.001]$ & -- \\
\bottomrule
\end{tabular}
\end{table}

The controller's advantage over the Base LM is unchanged under naturalized
language, and the Base LM degrades more than the controller does: exact-query
accuracy falls by .032 for the Base LM (significant on every
checkpoint) and by .006 for SAC-P (not significant). Because the user
simulator differs, these numbers are a separate control and are not
comparable row-for-row with Table~\ref{tab:control-main}. Database equivalence improves by $+.285$
$[+.275,+.294]$ over the Base LM under naturalized language, against $+.260$
with template users.

\section{Prompts and Dialogue Templates}
\label{app:qualitative}

This appendix gives the exact prompts, query serialization, and matched
dialogue template used in the MultiWOZ experiments.

\subsection{Exact MultiWOZ Prompts and Query Formatting}
\label{app:exact-prompts}

\paragraph{Base LM SEARCH.}
All final MultiWOZ analyses use the v2 SEARCH prompt. The rendered prompt is
the last 64 dialogue lines prefixed by \texttt{[DIALOGUE]}, followed by

\begin{quote}\ttfamily
[SEARCH]\\
Write the search constraints for the user's current request on one line as
slot: value pairs separated by semicolons, using only what the user has told
you. Begin with domain, chosen from hotel, restaurant, attraction. Then use
only these slot names: area, pricerange, food, type, stars, parking, internet,
name. Use the user's own words for the values.
\end{quote}

\paragraph{Base LM response.}
After database execution, the model receives the same rendered dialogue
window followed by

\begin{quote}\ttfamily
[DATABASE]\\
\{db\}\\[2mm]
[SYSTEM RESPONSE]
\end{quote}

where at most three returned entities are rendered into \texttt{\{db\}}.
DB-equivalence metrics
are computed from the complete returned entity set, not only the displayed
subset.

The natural-failure value scoring (Appendix~\ref{app:operational-profiles})
and slot forcing (Appendix~\ref{app:slot-forcing}) use the scaffolds
\texttt{domain: \{dom\}; \{slot\}:} and \texttt{; \{slot\}:} described there,
not the SEARCH prompt.

\subsubsection{Query Serialization}

Queries follow \texttt{domain: DOMAIN; slot: value; \ldots}, with the domain
and slot vocabulary of Appendix~\ref{app:closed-loop-protocol}. Parsing is insensitive to slot order and tolerates whitespace variation.
The deterministic controller compiler instead emits fields in the canonical
domain-specific constrained-slot order. Slot and value aliases are
canonicalized before execution; for example,
\texttt{location} maps to \texttt{area},
\texttt{cuisine} to \texttt{food}, and
\texttt{downtown} to \texttt{centre}.

A resolved value of \texttt{dontcare} is represented by omission of that slot,
and a domain-only query returns the unconstrained domain result set.

Importantly, syntactic parsing and semantic legality are distinct. For example,

\begin{quote}
\texttt{domain: restaurant; area: Cambridge;}
\end{quote}

parses successfully as
\texttt{\{"area":"cambridge"\}}, but the value is outside the restaurant-area
vocabulary and the database returns no entities.

DB equivalence requires the full returned entity-name set to equal that of
the gold query; empty result sets are never counted as DB-equivalent.

\subsection{Matched Historical-Source Construction}
\label{app:p4-example}

The matched-validity intervention in Appendix~\ref{app:matched-validity}
uses synthetic twin dialogues in which the historical source text and token
positions are held fixed while only the final update instruction changes.
These stimuli hold lexical content and position fixed to isolate source
validity; natural failures are analyzed in Section~\ref{sec:natural}.

A generic pair has the following form:

\begin{quote}
\textbf{User:} I am looking for a $v_1$ $d$.\\
\textbf{System:} I can help with that. What area of town do you prefer?\\
\textbf{User:} I would like it to be in the $a$.\\
\textbf{System:} Noted. Is there anything else you would like to change?\\
\textbf{User:} On second thought, make it $v_2$ instead.\\
\textbf{System:} Understood. I have made a note of that.
\end{quote}

Here $v_1$ and $v_2$ are values of the target slot, $d$ is the domain (restaurant or hotel), $a$ is an area value that forms the other-constraint span, and $s$ is the target slot's name (e.g., price range). The twins differ only in the final user instruction:

\begin{quote}
\textbf{Superseded-source twin:}
Yes, please apply that last change and use my new $s$.\\[1mm]
\textbf{Current-source twin:}
No, please ignore that last change and keep my first $s$.
\end{quote}

Thus the original $v_1$ source is lexically identical and occupies the same
token positions in both twins, but it is superseded in the first dialogue and
current in the second. This construction isolates how downstream influence of
the same historical source changes with conversational validity.

\end{document}